\documentclass[11pt,a4paper]{article}
\usepackage[utf8]{inputenc}
\usepackage[T1]{fontenc}
\usepackage{lmodern}
\usepackage[margin=2.5cm]{geometry}
\usepackage{graphicx}
\usepackage{amsmath,amssymb}
\usepackage{booktabs}
\usepackage[font=small,labelfont=bf]{caption}
\usepackage{etoolbox}
\AtBeginEnvironment{table}{\small}
\usepackage{authblk}
\usepackage[numbers,sort&compress]{natbib}
\usepackage[ruled,vlined,linesnumbered]{algorithm2e}
\usepackage{makecell}
\setcellgapes{3pt}
\usepackage[colorlinks=true,allcolors=blue]{hyperref}
\usepackage{orcidlink}
\title{When Hospital Robots Ask Staff for Help:\\ Forecasting Human Support in Robot Dispatch}
\author[1]{Krzysztof Siwek\,\orcidlink{0000-0003-2642-2319}\thanks{Corresponding author: \href{mailto:krzysztof.siwek@pw.edu.pl}{krzysztof.siwek@pw.edu.pl}}}
\author[2]{Aleksandra \'{S}wietlicka\,\orcidlink{0000-0002-2576-7625}\thanks{\href{mailto:aleksandra.swietlicka@put.poznan.pl}{aleksandra.swietlicka@put.poznan.pl}}}
\affil[1]{Faculty of Electrical Engineering, Warsaw University of Technology, Koszykowa 75, 00-662 Warsaw, Poland}
\affil[2]{Institute of Automatic Control and Robotics, Pozna\'n University of Technology, ul. Jana Paw{\l}a II 24, 61-138 Pozna\'n, Poland}
\date{}

\begin{document}
\maketitle

\begin{abstract}
Hospital delivery robots still need people to load, receive, and clean them, so every robot mission is also a request for help addressed to the staff it is meant to relieve. We ask whether a dispatcher that forecasts this support chooses better robot missions or merely fewer of them, and what its choices ask of staff. A discrete event simulation replays 33,079 recorded requests of a university hospital, with robot missions supported by mobile transporters, by stationary local workers, or by both. Forecasting support lowers weighted tardiness by 0.163 weighted minutes per request relative to the same rollout with omitted support. Against random admission at matched coverage this benefit is almost entirely a volume effect, but at the same coverage the rule with omitted support is 0.090 weighted minutes per request worse than the support-aware rule, because it sends urgent requests to robots while the transporters needed for their handovers are busy; this comparison is unchanged when urgent requests make up 20 instead of 40\% of demand. Seen from the staff side, a tardiness objective addresses most robot calls to busy people: 46\% of the missions of the support-aware rule start when no transporter is idle, against 18\% under random admission, and released staff time overstates the relief of transporters, whose other duties are displaced as much as under human transport. A gate that admits a robot only when idle workers can serve it halves the calls to staff and lowers tardiness by a further 0.086 weighted minutes per request, mostly by admitting fewer robots. A CP-SAT scheduler adds no selection value beyond the rollout, and human transport keeps the lowest mean tardiness. Dispatch rules for hospital robots should therefore be judged by what they ask of staff as well as by the service they deliver.
\end{abstract}


\section{Introduction}\label{sec:intro}
	Autonomous mobile robots now carry specimens, medication, linen, meals, and sterile instruments in many hospitals, and nursing managers expect them to relieve clinical staff of non-nursing work \cite{holland2021,kriegel2022}. Field studies show, however, that delivery robots rarely work alone. Someone has to put the specimen in the drawer, take the medication out at the ward, and wipe the cart after an infectious load, and staff perceive the robot as a colleague whose needs compete with their own work \cite{mutlu2008,ozkil2009,ljungblad2012}. Recent interview and observational studies with nursing staff point the same way: nurses regard the transport of materials as a suitable task for robots but expect to stay involved at several points of the process, and the walks and interruptions that service robots are meant to remove are tied to the stress that nurses report \cite{ohneberg2025,klebbe2025,brunett2025,warmbein2025}. Each robot mission borrows minutes from the same people who would otherwise carry the item by hand. A dispatcher deciding whether a request goes to a robot or to a transporter must weigh the travel a person saves against the support a person must still give, and against the delay that the robot's slower route and the waiting for help impose on other work.

	Seen from the side of human--robot interaction, every robot mission is a short sequence of encounters between a robot and the people who load, receive, and clean it. Whether these encounters fit into the work of staff depends not only on the robot's interface but on when they are created, and in a fleet they are created by the dispatcher: a mission admitted while every transporter is busy becomes a call that interrupts someone, or a robot that stands at a door until someone comes. Interruptions are a known hazard of clinical work \cite{westbrook2010}, alerts compete for the attention of nurses \cite{cvach2012}, and robots that need help are treated better when they choose whom to ask and when \cite{rosenthal2012,kimoto2022}. A dispatch rule is therefore also a policy for robot requests for help, and its evaluation should report what it asks of staff next to the service it delivers.

	A natural design response is to forecast support explicitly: before admitting a robot, simulate how loading, receiving, and cleaning will occupy workers and how this delays the rest of the queue. In the experiment reported here, such a forecast clearly reduces tardiness relative to one that ignores support. That comparison, however, mixes two mechanisms. A forecast that sees support costs will reject more robot missions, and if robots are on average worse than people, admitting fewer of them helps whatever the choice of missions. Only the residual, the improvement at a fixed number of robot missions, reflects better selection. The distinction matters for robotic system design: if the gain is a volume effect, a simple admission quota achieves it; if it is a selection effect, the dispatcher needs a model of the human resource.
	
	This paper separates the two mechanisms in two ways. Control arms borrowed from experimental design \cite{imbens2015} admit robots at random with the probability that reproduces, in every condition, the coverage of each forecast rule; and a threshold version of the forecast without support is calibrated to the coverage of the support-aware forecast, so that the two forecasts can be compared at the same number of robot missions. Mission logs then record which requests each rule gives to robots and when. We ask three questions.
	
	\begin{itemize}
		\item[RQ1] How much of the benefit of forecasting staff support is due to admitting fewer robots, how much to choosing better missions, and which missions do the forecasts choose?
		\item[RQ2] Does exact scheduling of the known support work, with a constraint programming solver, add selection value beyond a heuristic rollout?
		\item[RQ3] Under which operating changes, such as fewer transporters, a fixed staff budget, unattended receiving, higher demand, night-only operation, or multi-load tours, could robots compete with human transport?
		\item[RQ4] What do the rules ask of the staff who support the robots: how often are workers called, how long do robots wait for them, in which state of the workforce do the calls arrive, and which other duties do they displace? Can a rule keep its calls away from busy staff without losing service?
	\end{itemize}
	
	An earlier study by the same authors \cite{siwek2026arxiv} built a resource-aware dispatch controller on the same request data and compared it with human dispatch; the present paper asks where the benefit of such admission rules comes from. The contributions are as follows. (i) A volume and selection decomposition for robot admission rules along two paths, with random controls at the coverage of each compared rule, a random admission curve per condition that corrects imperfect calibration, and a comparison of competing rules at equal coverage. (ii) Evidence that random controls alone attribute the benefit of support forecasting entirely to volume, whereas at equal coverage more than half of it is selection, and mission logs showing why: the forecast without support gives its decisive robot missions to urgent requests while the transporters needed to support them are busy. Exact scheduling adds no selection value beyond that of the rollout. (iii) A front of tardiness against released staff time, with random arms along it, showing that support forecasts select better at every coverage examined and buy staff time more cheaply. (iv) A set of operating scenarios showing that the staff time per mission, rather than transport capacity or dispatch logic, limits robot benefit in this model, and a check that the selection results do not depend on the assumed share of urgent requests. (v) A staff-side evaluation of the rules, showing that a tardiness objective addresses most robot calls to transporters who are all busy and that released staff time overstates their relief, and an availability gate that keeps calls away from busy staff, halves them, and improves service. All results are conditional on a single hospital pathway and assumed service parameters.
	
	The paper is organised as follows: Section~\ref{sec:related} reviews related work. Section~\ref{sec:model} describes the simulation model and Section~\ref{sec:policies} the dispatch policies. Section~\ref{sec:design} presents the experimental design and the decomposition, Section~\ref{sec:results} the results, and Section~\ref{sec:discussion} their implications and limitations. Section~\ref{sec:conclusion} concludes.
	
	\section{Related work}\label{sec:related}
	\subsection{Hospital logistics robots and the work around them}
	Early deployments of hospital delivery robots showed that their value depends on how they fit existing workflows. Mutlu and Forlizzi \cite{mutlu2008} found that the same robot was welcomed in one hospital unit and resented in another, depending on workload and the interruptions it caused. \"Ozkil et al.\ \cite{ozkil2009} characterized the transport tasks of a Danish hospital as candidates for automation, and Ljungblad et al.\ \cite{ljungblad2012} described how staff alternately treated a delivery robot as an alien object, a machine, a worker, and a work partner. Recent reviews and surveys document a rapid expansion of service robots in healthcare \cite{holland2021} and show that nursing officers prioritize relief from non-nursing tasks \cite{kriegel2022}. Operational studies have evaluated autonomous mobile robots in sterile instrument logistics \cite{fragapane2023}, simulation-based control of automated guided vehicles \cite{lopez2022}, stochastic scheduling of hospital robots with time windows \cite{cheng2023}, and fleet coordination with shared elevators and batches \cite{vandonkelaar2026}. Gombolay et al.\ \cite{gombolay2018} showed that a robot can support the resource nurse who coordinates patient care, which underlines that hospital robots operate inside human decision processes. Studies in the \emph{International Journal of Social Robotics} have examined how nursing staff perceive such systems and what their integration requires. In interviews, nurses name the transport of materials, meals, and beverages among the useful applications of robotic assistance in acute care, but they also set spatial, personnel, organisational, and technical conditions for its integration \cite{ohneberg2025}. Caregivers grant robots a high degree of autonomy for suitable tasks, yet expect to be involved at various times so that they keep their responsibility for the care process \cite{klebbe2025}, and focus groups stress that robot use must be adapted to work processes and to the condition of patients \cite{brunett2025}. In a seven-month field study on a nuclear medicine ward, a delivery robot combined with a communication app had no direct effect on the walking distances or stress of nurses, while the number of walks was associated with perceived interruption and stress \cite{warmbein2025}. In these studies, staff appear as users, supervisors, or fixed handling times; the time they spend supporting robots is rarely modeled as a finite resource that robots compete for.

	\subsection{Robots that ask people for help}
	A delivery robot that needs a person to load and receive it is a robot that asks for help, and studies of such robots treat the request as a social act whose timing and addressee matter. Rosenthal et al.\ \cite{rosenthal2012} let an office robot decide whom to ask from estimates of the occupants' availability, Srinivasan and Takayama \cite{srinivasan2016} found that the politeness strategy of a robot's request changed people's willingness to help, and Kimoto et al.\ \cite{kimoto2022} timed robot requests by estimating the subjective time of the person to be asked. Interruptions in general carry costs that depend on their timing and on the task they disrupt \cite{mcfarlane2002}; on hospital wards, interruptions of nurses are associated with medication administration errors \cite{westbrook2010}, and the volume of device alarms produces alarm fatigue \cite{cvach2012}. At the level of teams, the share of decision authority that a robot holds in task allocation changes both team efficiency and the satisfaction of workers \cite{gombolay2015}, and Hoffman \cite{hoffman2019} proposed fluency measures for human--robot collaboration, among them the idle time of the robot and of the human and the delay between their actions. A long-term deployment of an autonomous robot in a care hospital met mixed acceptance among staff \cite{hebesberger2017}, and acceptance of robots at work varies between workers and workplaces \cite{turja2019}; trust in automation, finally, depends on whether its behavior is predictable to the people who rely on it \cite{lee2004}. These studies examine single requests, single robots, or small teams. In a hospital fleet, the number and timing of robot requests for help are set by the dispatch rule, and how a rule distributes them over busy and quiet periods has, to our knowledge, not been measured.

	\subsection{Task allocation with human resources}
	Multi-robot task allocation is usually framed as an assignment under resource and timing constraints \cite{gerkey2004,korsah2013}. In the iTax taxonomy of Korsah et al.\ \cite{korsah2013}, our problem has cross-schedule dependencies: the value of assigning a delivery to a robot depends on the schedules of the people who must load and receive it. Temporal and ordering constraints between such tasks have been classified by Nunes et al.\ \cite{nunes2017}. For healthcare fleets, distributed auctions \cite{das2015} and auction-based allocation under human supervision \cite{galati2023} have been proposed. On the interaction side, the physical exchange of objects between people and robots has been studied as a handover problem \cite{ortenzi2021}, including how a robot should behave while it receives an object from a person \cite{faibish2022}; we abstract each loading and receiving step to a service time that requires an available worker at the right place.
	
	\subsection{Dynamic transport dispatching and rollout}
	Requests that arrive during operation make robot dispatch a dynamic pickup and delivery problem \cite{berbeglia2010,pillac2013}, and intralogistics adds routing, charging, and shared infrastructure \cite{fragapaneplanning2021,leanh2006}. In hospitals, the closest operational analog is the dispatching of patient transports: Hanne et al.\ \cite{hanne2009} and Beaudry et al.\ \cite{beaudry2010} developed planning and dynamic dial-a-ride heuristics for patient transport services, and Kergosien et al.\ \cite{kergosien2011} added urgent priorities and disinfection after infectious transports. Our forecast rules are one-step rollouts: each candidate action is scored by simulating a deterministic continuation under a base policy \cite{bertsekas1997,powell2011}. Rollout is well established for dynamic routing with stochastic requests \cite{goodson2017,ulmer2019}; its quality depends on the continuation model, which is why we compare it with rules that use no forecast at all and with an exact scheduler \cite{perron2023}.
	
	\subsection{Evaluation methodology}
	We follow the STRESS guidelines for reporting simulation studies \cite{monks2019}, distinguish verification from validation \cite{sargent2013}, adjust primary contrasts with the Holm procedure \cite{holm1979}, and note that confidence intervals from paired replications do not include input uncertainty \cite{barton2014}. The random arm at matched coverage plays the role of a placebo control in a randomized comparison \cite{imbens2015}: it receives the same amount of the intervention, robot missions, without the mechanism under test. To our knowledge, such volume-matched controls have not been used to evaluate robot dispatch rules.
	
	\section{Simulation model}\label{sec:model}
	\subsection{Demand and physical execution}
	We use a public dataset of the University Medical Center Freiburg \cite{hess2026}, containing transport requests (33,079), endpoints (1,913) and paths (6,707 nodes and 13,834 edges). The physical transport model, with shared elevators, cleaning stations, and chargers, is taken from \cite{siwek2026arxiv}, and we extend it with support arrangements, reporting delay, forecast-based admission rules, and randomized controls. Each simulated day is drawn with replacement from the 92 recorded days, which preserves the request times and endpoints of a real day (360 requests on average). Routes follow the observed paths where these are available and are otherwise estimated from the graph. Because the records carry no task type, each request receives a class drawn from the assumed distribution of Table~\ref{tab:tasks}; specimens are treated as urgent and weighted $w_j=4$ in the tardiness measure, whereas all other classes have weight 1.
	
	\makegapedcells
	\begin{table}[!htbp]
		\caption{Assumed task classes. Times in minutes. Priority orders the human job queue (1 is served first); the weight enters the tardiness measure $J$. Loading and receiving apply to both human and robot deliveries.}\label{tab:tasks}
		\centering
		\begin{tabular}{lrrrrrr}
			\toprule
			Class & Share (\%) & Priority & Weight $w_j$ & Deadline & Loading & Receiving \\
			\midrule
			Specimen & 40 & 1 & 4 & 25 & 1.0 & 0.8 \\
			Medication & 30 & 2 & 1 & 35 & 1.4 & 1.0 \\
			Supply & 20 & 4 & 1 & 120 & 2.0 & 1.5 \\
			Instrument & 10 & 2 & 1 & 45 & 2.5 & 1.5 \\
			\bottomrule
		\end{tabular}
	\end{table}
	The mobile workforce consists of 8, 30 and 24 transporters working during the night, day and evening, respectively, giving 496 staffed hours per day. People walk at 72~m/min and the two robots drive at 54~m/min, and every travel time is multiplied by a congestion factor drawn from a lognormal distribution with mean 1 and log-standard deviation 0.18, truncated below at 0.65. People and robots share one bank of four elevator cars served first come, first served, with a cycle of 2.2 minutes plus 0.25 minutes per floor.
	
\begin{figure}[!htbp]
		\centering
		\includegraphics[width=\textwidth]{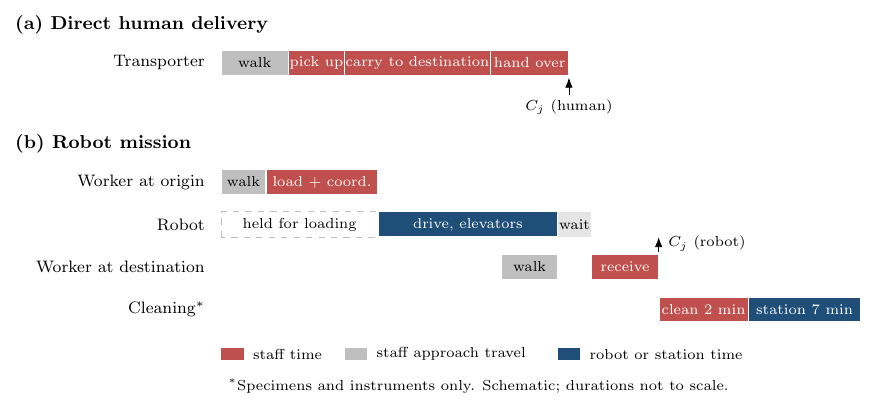}
		\caption{Staff touchpoints of (a) a direct human delivery and (b) a robot mission. Red segments occupy a worker, grey segments represent worker travel, and blue segments occupy only the robot or a cleaning station. The robot saves the carrying time but adds a second worker approach, coordination, waiting, and, for specimens and instruments, cleaning. Released staff time is the manual delivery time avoided minus all red and grey segments of the mission.}\label{fig:timeline}
	\end{figure}

	Figure~\ref{fig:timeline} contrasts the two ways of fulfilling a request. In a \emph{human delivery}, the transporter walks to the origin, loads the item (Table~\ref{tab:tasks}), carries it, and hands it over at the destination. In a \emph{robot mission}, a worker walks to the robot at the origin and loads it, spending in addition a coordination time of $2e^{Z}$ minutes with $Z\sim N(0,0.3^2)$, which a robot fault (probability 0.02) extends by an exponential recovery time with mean 12 minutes; after the robot has driven to the destination, a worker must come there to receive the item. Specimen and instrument missions then require 2 minutes of cleaning by staff and 7 minutes at one of the two cleaning stations. Each robot has a 4~kWh battery, consumes 0.025~kWh per active minute, and recharges at one of two 2~kW chargers once its charge falls below 25\%. Under human transport alone, transporters spend 18 to 21\% of their staffed time on deliveries, so that time remains for the other duties represented by the unavailability process of Section~\ref{sec:arr}; the main experiment thus represents a hospital in which transport is not a bottleneck.
	
	\subsection{Support arrangements and unavailability}\label{sec:arr}
	Three support arrangements are compared. In the \emph{mobile} arrangement, transporters perform direct deliveries and all robot support. The \emph{local} arrangement adds one stationary worker at each of the 95 locations, who loads, receives, and cleans robots arriving there. It represents ward or laboratory staff whose primary duties are modeled by the unavailability process below. The \emph{mixed} arrangement uses local workers for receiving only. Local workers add 2,280 staffed hours per day, so comparisons across arrangements change capacity; all policy comparisons are made within an arrangement, and released staff time counts the support minutes of both mobile and local workers.
	
	Each worker follows a 30-minute cycle with a random offset and, during a fraction $b\in\{0,0.4,0.8\}$ of each cycle, accepts no new assignment while finishing the current one. The resulting calendar is drawn from a random stream that does not depend on the policy. Human jobs are ordered by priority, due date, and arrival, and each is assigned to the nearest available eligible worker.
	
	\subsection{Information}
	
	The dispatcher has knowledge of the queue, shifts, robot states, and resource usage. Worker locations and unavailability are reported either immediately or with a delay of 10 minutes. A busy worker is predicted to become free at the nominal completion time of the current assignment, and a worker reported as unavailable is expected back after $15b$ minutes. The forecasts consider neither future requests nor future unavailability periods, and they draw no random numbers.
	
	\subsection{Verification}
	
	Fourteen automated tests verify the simulator \cite{sargent2013}. They check that requests and worker time are conserved in the human baseline, that the mobile baseline is reproduced when no worker is unavailable, and that local support never draws on mobile workers or on workers from other locations, so that a robot waits when its local worker is busy. Further tests confirm that the unavailability calendar is reproducible and independent of the policy, that forecasts draw no random numbers, that delayed reports never reveal current worker locations, and that shift constraints are respected. The optimizer matches exhaustive enumeration on single-worker instances and raises model errors instead of silently falling back on the heuristic, and random admission with probability zero reproduces human transport exactly. Fourteen instead of seven warm-up days change the tardiness of human transport and staff rollout by $-0.2$ to 6.9\% in three conditions with $b=0.4$ and 0.8, and none of these changes differs from zero ($p\ge0.53$, 20 seeds, unpaired).
	
	\section{Dispatch policies}\label{sec:policies}
\begin{figure}[!htbp]
		\centering
		\includegraphics[width=0.92\textwidth]{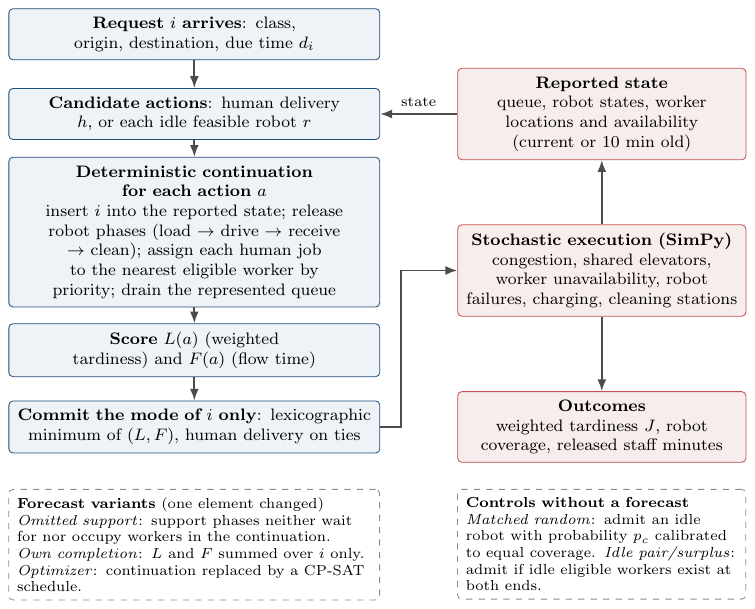}
		\caption{Decision flow. At each request, the dispatcher scores every candidate action by a deterministic continuation of the reported state and commits only to the transport mode of the arriving request; execution then proceeds in the stochastic simulator. Dashed boxes list the variants and controls compared in this study.}\label{fig:flow}
	\end{figure}

	Figure~\ref{fig:flow} summarizes how the dispatcher and the simulator interact, and Table~\ref{tab:notation} lists the policies and the main quantities.
	
	\begin{table}[!htbp]
		\caption{Policies and main quantities.}\label{tab:notation}
		\centering
		\begin{tabular}{p{0.27\textwidth}p{0.65\textwidth}}
			\toprule
			Policy or symbol & Meaning \\
			\midrule
			Human transport & All requests delivered by transporters; baseline \\
			Staff rollout & One-step rollout whose continuation represents the occupation of workers by robot support (Algorithm~\ref{alg:rollout}) \\
			Omitted support & Same rollout, but support phases neither wait for nor occupy workers in the continuation \\
			Own completion & Staff rollout with $L$ and $F$ summed over the arriving request only \\
			Threshold rule & Omitted support, admitting only if the predicted margin passes a threshold calibrated to the coverage of staff rollout \\
			Optimizer & Candidate actions scored by a CP-SAT schedule of all represented work \\
			Matched random & Idle robot admitted with probability $p_c$ calibrated to the coverage of a reference rule \\
			Idle pair / surplus rule & Robot admitted if idle eligible workers exist at both ends (surplus: and the queue is empty, with two idle mobile workers per phase) \\
			Availability gate & Staff rollout, admitting only if idle eligible workers exist for loading and receiving \\
			Allowance rule, $\varphi$ & Robot admitted at equal predicted tardiness if it saves staff minutes and adds at most $\varphi$ minutes of flow time \\
			\midrule
			$b$ & Nominal fraction of time a worker is unavailable \\
			Condition $c$ & Arrangement $\times$ $b$ $\times$ reporting delay (18 in the main experiment) \\
			$J$ & Weighted tardiness per measured request, weighted minutes \\
			Coverage & Share of requests completed by robots \\
			Released time & Manual transport time avoided minus staff time spent on robot support, min/day \\
			$B$, $V$, $S$ & Benefit of representing support, its volume and selection terms (Eq.~\ref{eq:decomp}) \\
			$V^\ast$, $S_{\mathrm{omit}}$ & Pure volume effect and selection of omitted support at its own coverage (Eq.~\ref{eq:three}) \\
			\bottomrule
		\end{tabular}
	\end{table}
	
	\subsection{Forecast rules}
	At the arrival of request $i$, the candidate actions are direct delivery and all idle robots. The deterministic roll-out policies corresponding to each action add the request to the state, release phases from robots in order, allocate all jobs to the closest worker (according to priority), and empty the queue. This includes at most 32 delivery requests, all support requests, and the rest of the phases of active missions. Using the predicted completion times $\widehat C_j(a)$, the due dates $d_j$, arrival times $r_j$ and weights $w_j$, the \emph{staff rollout} selects the lexicographical minimum of
	
	\begin{equation}\begin{split}
			L(a)&=\sum_{j}w_j\,[\widehat C_j(a)-d_j]_+,\\ F(a)&=\sum_j\bigl(\widehat C_j(a)-r_j\bigr),
		\end{split}
		\label{eq:score}
	\end{equation}
	keeping the human option on ties (Algorithm~\ref{alg:rollout}). Only the transport mode of request $i$ is committed. Because a robot is slower than a person, $F$ usually favors human delivery when no request is predicted late, so the rule admits a robot mainly when it lowers predicted tardiness of the queue.
	
	\begin{algorithm}[!ht]
		\caption{Staff rollout decision at the arrival of request $i$}\label{alg:rollout}
		\KwIn{reported state $s$ (queue, robots, workers), request $i$, idle feasible robots $\mathcal R$}
		\KwOut{transport mode of $i$}
		\ForEach{$a\in\{h\}\cup\mathcal R$}{
			$\hat s\leftarrow$ copy of $s$ with $i$ inserted in mode $a$\;
			\While{$\hat s$ has unassigned represented jobs}{
				release the next robot phase (load, drive, receive, clean) whose predecessor is complete\;
				assign the highest-priority human job (delivery or support) to the nearest eligible worker predicted free\;
				advance predicted completion times $\widehat C_j(a)$\;
			}
			compute $L(a)$ and $F(a)$ by Eq.~\eqref{eq:score}\;
		}
		$a^\ast\leftarrow\arg\min_a \bigl(L(a),F(a)\bigr)$ lexicographically, with $h$ preferred on ties\;
		\Return{$a^\ast$}
	\end{algorithm}
	
	Two variants change only the forecast. \emph{Omitted support} removes from the continuation the waiting for and occupation of workers during loading, receiving, and cleaning, although these phases are still executed in the simulation; \emph{own completion} restricts the sums of Eq.~\eqref{eq:score} to request $i$. All variants share the workforce and the dispatching rules for human jobs.
	
	\subsection{Controls without a forecast}
	
	\emph{Matched random} admits an idle feasible robot with probability $p_c$ in condition $c$, drawing the decision from a hash of the request so that the random streams of the simulator are unaffected. Each $p_c$ was calibrated on separate pilot seeds (104900 to 104907) with two warm-up and four measured days, starting from unrestricted admission and applying two multiplicative corrections. The \emph{idle pair rule} admits the nearest robot if its mission can meet the deadline and idle eligible workers are available for loading and receiving; the \emph{idle surplus rule} further requires an empty queue and two idle mobile workers for each phase. Both rules represent a dispatcher who observes the workforce but does not forecast the queue.
	
	\subsection{Staff time allowance}
	To trace the trade off between tardiness and staff time, a family of rules admits robot $r$ over human delivery $h$ if $L(r)<L(h)$, or if $L(r)=L(h)$, the predicted staff minutes $H(r)<H(h)$, and $F(r)-F(h)\le\varphi$. The allowance $\varphi\in\{3,10\}$ minutes of flow time is the delay accepted for a predicted staff saving. The rule is applied to the staff continuation and to the continuation without support, where every robot appears to save the full delivery time.

	\subsection{Availability gate}\label{sec:gate}
	The \emph{availability gate} keeps the continuation and the choice of staff rollout but admits the chosen robot only if, in the reported state at the decision, an idle worker is eligible for loading and an idle worker is eligible for receiving; this is the worker condition of the idle pair rule without its deadline test. The gate represents a dispatcher that does not call staff who are all busy, as a person coordinating the robots might do. With the condition disabled, the implementation reproduces staff rollout bit for bit. The rule was added after the staff-side analysis of Section~\ref{sec:staff} had shown where the calls of staff rollout arrive.
	
	\subsection{Bounded optimization}
	The optimizer evaluates the same candidate actions by solving, for each action, a scheduling problem over all represented human deliveries and robot support phases using the CP-SAT solver in OR-Tools \cite{perron2023,ortools}. Each operation is assigned to one eligible worker; a circuit over each worker's operations enforces approach travel, release times, shift starts, and precedence between loading, receiving, and cleaning. The solver first minimizes weighted tardiness and then, at that optimum, total completion time. It runs with one search worker and a fixed random seed and receives one second per solve, with retries of 5 and 20 seconds only after an unknown status. The heuristic continuation is used only if no schedule is found. During execution, the optimizer also reorders the human queue. Its comparison with staff rollout evaluates admission and scheduling together.
	
	\section{Experimental design and analysis}\label{sec:design}
	\subsection{Main experiment and outcomes}
	
	The main experiment crosses three support arrangements, three values of $b$, and two reporting delays, giving 18 conditions, and runs every policy on 20 seeds (103000 to 103019) that drive the demand and unavailability streams. Each run comprises seven warm-up days, seven measured days, and one clearance day. For each seed, the difference between two policies is averaged with equal weight over the conditions, and 95\% Student $t$ confidence intervals of the mean are computed across seeds; they are given as (lower to upper) in the text and as [lower, upper] in the tables. Because tardiness at $b=0.8$ is about three times that at $b=0$, mean absolute differences are dominated by high unavailability, so results are also reported by subset and as relative differences. The three main contrasts (staff rollout against omitted support, own completion, and the optimizer) are adjusted with the Holm procedure \cite{holm1979}. The contrasts with matched random and the idle pair rule are reported with Holm $p$ values as secondary tests; all other contrasts are descriptive.

	The weighted tardiness is calculated as:
	\begin{equation}
		J=\frac{1}{|\mathcal M|}\sum_{j\in\mathcal M} w_j\,\left\lbrack C_j-d_j\right\rbrack_+ ,
	\end{equation}
	(in weighted minutes). All measurement requests were completed by the end of the clearance day. The proportion late is the proportion of measurement requests where $C_j>d_j$. Robot coverage is the proportion of requests completed by robots. Released staff time is the amount of manual transport time saved minus the amount of time staff spend approaching, loading, coordinating, receiving, and cleaning robots (in minutes per day).

	Released time is an account of the robot system, not of the experience of the people in it. To describe what a rule asks of staff, every run also records the following staff-side measures. A \emph{call} is one loading, receiving, or cleaning assignment of a worker to a robot; the \emph{call-to-arrival time} runs from the moment the robot needs the worker to the start of the service and comprises the wait for a free worker and the approach walk, so that, summed over a day, it is the time robots stand waiting for people, the robot idle time of the fluency measures of Hoffman \cite{hoffman2019}. \emph{Displaced other duties} are the minutes of the unavailability periods of Section~\ref{sec:arr} that a worker could not take up because a delivery or support job was still running, the planned minus the performed minutes of these periods, and \emph{overtime} counts the minutes of jobs that run past the end of a shift. The mission logs (Section~\ref{sec:vsdesign}) add the number of idle transporters at each admission. All of these are quantities of the model; they describe the load that the rules place on staff, not the workload or interruption that staff perceive.

	\subsection{Volume and selection decomposition}\label{sec:vsdesign}
	The benefit of representing support, $B=J_{\mathrm{omit}}-J_{\mathrm{staff}}$, splits additively into
	\begin{equation}
		B=\underbrace{J_{\mathrm{omit}}-J_{\mathrm{rand}}}_{V}+\underbrace{J_{\mathrm{rand}}-J_{\mathrm{staff}}}_{S},
		\label{eq:decomp}
	\end{equation}
	where $J_{\mathrm{rand}}$ is matched random calibrated to the coverage of staff rollout. $S$ is the value of choosing which requests robots serve at equal coverage. $V$ compares omitted support with random admission at the coverage of staff rollout. It contains the effect of the different number of robots and, in addition, any difference between the selection of omitted support and random selection. A split that isolates the volume effect needs a second random arm at the coverage of omitted support, $R_{\mathrm{omit}}$:
	\begin{equation}
		B=\underbrace{(J_{\mathrm{omit}}-J_{R_{\mathrm{omit}}})}_{-S_{\mathrm{omit}}}+\underbrace{(J_{R_{\mathrm{omit}}}-J_{\mathrm{rand}})}_{V^\ast}+S.
		\label{eq:three}
	\end{equation}
	Here $V^\ast$ is the pure volume effect, random admission at the coverage of omitted support against random admission at the coverage of staff rollout, and $S_{\mathrm{omit}}=J_{R_{\mathrm{omit}}}-J_{\mathrm{omit}}$ is the selection value of omitted support at its own coverage. As in other sequential decompositions, such as the Oaxaca--Blinder decomposition, the split depends on the path: $S$ and $S_{\mathrm{omit}}$ are measured at different coverages, so their difference mixes selection quality with the number of decisions in which selection can act. The arm $R_{\mathrm{omit}}$ was calibrated per condition to the mean coverage of omitted support with the same pilot procedure as matched random (Section~\ref{sec:policies}) and run on all main seeds and conditions.
	
	The second path compares the two forecasts directly at equal coverage. The \emph{threshold rule} keeps the continuation and the lexicographic choice of omitted support but admits the best robot only if its predicted margin over human delivery passes a threshold: first a minimum reduction of flow time $F$ for decisions in which $L$ is tied, then, for higher thresholds, a minimum reduction of $L$ with no tied decisions admitted. A threshold of zero reproduces omitted support exactly. The threshold was calibrated per condition on the pilot seeds to the coverage of staff rollout, so that
	\begin{equation}
		B=\underbrace{(J_{\mathrm{omit}}-J_{\mathrm{thr}})}_{\substack{\text{volume,}\\ \text{ranking of omitted support}}}+\underbrace{(J_{\mathrm{thr}}-J_{\mathrm{staff}})}_{\text{selection at equal coverage}} .
		\label{eq:thr}
	\end{equation}
	The first term removes the robot missions that omitted support values least; the second compares the missions that each forecast values most at the same number of missions.
	
	Calibration matches coverage only approximately. Instead of a single slope, each condition has a \emph{random admission curve}: random admission with probability zero, which reproduces human transport exactly, and the random arms at the coverage of staff rollout, omitted support, and the four allowance rules of Section~\ref{sec:policies} give seven points of $J$ against mean coverage. For a policy $P$ with mean coverage $\mathrm{cov}_P$ in a condition, the curve-adjusted selection value is $S_P=R(\mathrm{cov}_P)-J_P$, where $R$ interpolates linearly between the neighboring arms on the same seed. With the curve in place of $J_{\mathrm{rand}}$ and $J_{R_{\mathrm{omit}}}$, the identity of Eq.~\eqref{eq:three} remains exact.
	
	To identify what the forecasts select, staff rollout, omitted support, and matched random were rerun on the same seeds and conditions with a subclass of the simulator that only reads state. At each admission decision it records the length of the human queue and the numbers of idle mobile workers, local workers, and robots; after the run it summarises every measured robot mission by class, deadline, deterministic route duration and distance, elevator rides, shift, and weighted tardiness. The $R_{\mathrm{omit}}$ runs carry the same log. The reruns reproduced $J$, coverage, late share, released time, and prediction error of all 1,080 stored runs bit for bit. Finally, the weighted tardiness of every run is split into the part incurred by robot-served requests (\emph{own}) and by human-served requests (\emph{other}), so that each selection value can be attributed to the robot missions themselves or to their effect on other work.
	
	The same identity, with random arms calibrated to the coverage of the optimizer ($R_o$) and of staff rollout ($R_r$), gives:
	\begin{equation}\begin{split}
			J_{\mathrm{staff}}-J_{\mathrm{opt}}&=V_o+S_o-S_r,\\ V_o&=J_{R_r}-J_{R_o},\\
			S_o&=J_{R_o}-J_{\mathrm{opt}},\\ S_r&=J_{R_r}-J_{\mathrm{staff}}. 
		\end{split}
		\label{eq:opt}
	\end{equation}
	
	The difference between the arms $R_o$ and $R_r$ also estimates the mean slope $\beta$ of $J$ in coverage. Since the scenario phases have no random admission curve, their coverage-adjusted selection value is $S_{\mathrm{adj}} = S - \beta (\mathrm{cov}_{\mathrm{rand}} - \mathrm{cov}_{\mathrm{staff}})$.
	
	\subsection{Scenario phases}
	The control, front volume, equal coverage, scenario, class mix, and availability gate phases were designed after the main results had been obtained, each with its own frozen design file, calibration log, and code hashes (Table~\ref{tab:phases}). The scenario phases use two robots, mobile and local support with $b\in\{0.4,0.8\}$, both reporting delays, and seeds 103000 to 103011; each runs human transport, staff rollout, and omitted support, and matched random is recalibrated for every scenario and condition. \emph{Scarcity} reduces the roster to 53 or 44 transporters. \emph{Equal budget} adds $k\in\{3,6\}$ local workers on duty around the clock at the service points with the most requests (both ends for local support, destinations for mixed support) and reduces the roster to 53 and 44 transporters, respectively, so that staffed hours remain at 496 per day; these local workers can support 6\% and 16\% of requests under local support and receive 23\% and 36\% under mixed support. Under \emph{unattended receiving}, the robot deposits its load in a receptacle at the destination, with the same service time but without a worker. Since human transport saturates at a demand factor of about 1.25, the \emph{demand} phase scales demand by 1.1 and 1.2. \emph{Night only} restricts robot admission to the night shift, staffed by eight or four transporters. In \emph{multi-load tours}, a request that arises at the origin of a robot that has not yet departed joins its tour without a policy decision, provided that the robot has room for at most three loads, the predicted arrival meets the deadline of the new request, no load aboard becomes late, and the battery suffices; each load still needs its own loading and receiving. A second variant holds the robot at the origin for up to 10 minutes for further loads, as long as every load aboard stays forecast on time. The \emph{combined} scenario joins demand $\times1.2$, unattended receiving, and tours with holding. The \emph{class mix} phase tests the assumption that 40\% of requests are urgent specimens. It halves this share to 20\%, keeps the ratio 3:2:1 of medication, supply, and instrument requests (40.0, 26.7, and 13.3\%), and runs human transport, staff rollout, omitted support, matched random, and the threshold rule on the 18 conditions of the main experiment and seeds 103000 to 103009; matched random and the threshold rule were recalibrated to the coverage of staff rollout in the new mix with the pilot procedures of Section~\ref{sec:policies}, the threshold on a grid of eleven values. The \emph{availability gate} phase runs the rule of Section~\ref{sec:gate} on the main design.
	\begin{table}[!htbp]
		\caption{Experimental phases. Conditions: M, mobile; L, local; X, mixed support. Scenario phases use M and L with $b\in\{0.4,0.8\}$ and both delays.}\label{tab:phases}
		\centering
		\setlength{\tabcolsep}{4pt}
		\begin{tabular}{lllr}
			\toprule
			Phase & Content & Seeds & Runs \\
			\midrule
			Main & 5 policies, 18 conditions & 20 & 1,800 \\
			Controls & matched random, idle pair and idle surplus rules & 20 & 1,080 \\
			Staff time front & allowance rules, $\varphi\in\{3,10\}$ & 20 & 1,440 \\
			Front volume & random arms at the coverage of the four allowance rules & 20 & 1,440 \\
			Equal coverage & threshold rule at the coverage of staff rollout & 20 & 360 \\
			Sensitivity & 4 and 8 robots, robot at 72 m/min, support $\times0.5$, $\times1.5$ & 6 & 816 \\
			Optimizer volume & random arms at optimizer and rollout coverage & 20 / 6 & 552 \\
			Omitted volume & random arm at omitted-support coverage; logged reruns & 20 & 1,440 \\
			Scarcity, unattended & 53 or 44 transporters; receptacle at destination & 12 & 1,152 \\
			Equal budget & 3 or 6 local workers within 496 staffed hours (L, X) & 12 & 768 \\
			Operating levers & demand $\times1.1$, $\times1.2$; night only; tours; combined & 12 & 3,072 \\
			Class mix & 20\% specimens; 5 policies, 18 conditions & 10 & 900 \\
			Availability gate & gate on staff rollout, 18 conditions & 20 & 360 \\
			\bottomrule
		\end{tabular}
	\end{table}
	
	\subsection{Implementation}
	In total the analysis uses 15,180 runs of seven measured days (Table~\ref{tab:phases}), of which 1,080 are logged reruns; all runs of the front, equal coverage, class mix, and availability gate phases also log robot missions. The class mix and availability gate phases were added after all other phases and run on the same server as the other phases; the staff-side measures were computed from the stored runs of all phases. The simulator is written in Python 3.12 with SimPy 4.1.2, and the optimizer uses OR-Tools 9.15 \cite{ortools}; runs were executed in parallel single-threaded processes on a Linux server with an AMD Ryzen 9 7900X processor (12 cores), and the later phases reproduced the stored outcomes of earlier runs exactly. The published archive holds 24,661 run files: the 15,180 analysed runs, the 121 runs of the warm-up check and the zero-probability random arm (Section~\ref{sec:model}), and 9,360 pilot, calibration, verification, test, and probe runs on seeds 101000 to 104907, which were never used for inference. The reporting follows the STRESS-DES checklist \cite{monks2019}; Appendix~B of the online appendix in the code repository\footnote{\url{https://github.com/siwekk/hospital-robot-staff-support/tree/master/appendix}; Appendix~A holds the additional result tables A1--A5, Appendix~B the STRESS-DES checklist.} maps its items to the sections of this article.
	
	\section{Results}\label{sec:results}
	\subsection{Main comparisons}
	Table~\ref{tab:main} summarizes all policies of the main experiment. Staff rollout improves on omitted support by 0.163 weighted minutes per request (13\% of the comparator), with all 20 seeds favorable. Scoring the whole queue adds nothing demonstrable over scoring only the arriving request (0.012, interval crossing zero). The optimizer is 0.045 better than staff rollout (4.2\%). Human transport has the lowest average tardiness, 0.113 below staff rollout, or 8.2\% (5.9 to 10.5\%) in relative terms. The forecast rules dispatch robots rarely: staff rollout serves 0.74\% of requests, about 2.6 missions per day for two robots, and releases 11.9 staff minutes per day, against 6.6 missions for omitted support. The idle worker rules admit about 12\% of requests, release about 200 staff minutes per day, and are worse than human transport by 0.75 and 0.84, almost doubling its tardiness.
	
	\begin{table}[!htbp]
		\caption{Main experiment, 20 seeds and 18 conditions. Released: net staff minutes per day. Difference: the policy minus staff rollout, so positive values mean that the policy has higher tardiness than staff rollout; 95\% intervals; desc., descriptive.}\label{tab:main}
		\centering
		\setlength{\tabcolsep}{2.9pt}
		\begin{tabular}{lrrrrrr}
			\toprule
			Policy & $J$ & Late (\%) & Coverage (\%) & Released & Difference [95\% CI] & Holm $p$ \\
			\midrule
			Human transport & 0.968 & 4.98 & 0 & 0 & $-0.113$ $[-0.134,-0.093]$ & desc. \\
			Optimization & 1.036 & 4.95 & 0.52 & 8.4 & $-0.045$ $[-0.073,-0.017]$ & 0.0068 \\
			Own completion & 1.069 & 5.10 & 0.66 & 11.0 & $-0.012$ $[-0.027,0.003]$ & 0.106 \\
			Staff rollout & 1.081 & 5.16 & 0.74 & 11.9 & & \\
			Matched random & 1.121 & 5.25 & 0.77 & 14.2 & 0.040 $[0.014,0.067]$ & 0.0042 \\
			Omitted support & 1.244 & 5.62 & 1.89 & 31.2 & 0.163 $[0.148,0.178]$ & $<10^{-14}$ \\
			Idle surplus rule & 1.713 & 8.07 & 11.52 & 198.4 & 0.632 $[0.593,0.671]$ & desc. \\
			Idle pair rule & 1.804 & 8.30 & 11.85 & 204.7 & 0.724 $[0.691,0.756]$ & $<10^{-19}$ \\
			\bottomrule
		\end{tabular}
	\end{table}

	The service cost of the robots is not uniform (Table~\ref{tab:decomp}). It arises at $b=0.8$, where staff rollout is 0.332 (0.267 to 0.398), or 22.7\%, worse than human transport. At $b=0$ and $b=0.4$ the difference is not distinguishable from zero ($-0.003$, $-0.025$ to 0.020, and 0.010, $-0.015$ to 0.036), and the same holds with ten minute reporting delays (0.018, $-0.004$ to 0.039). By support arrangement, staff rollout is worse than human transport by 0.083 (0.052 to 0.114) with mobile, 0.099 (0.074 to 0.123) with mixed, and 0.158 (0.128 to 0.187) with local support. In one condition, local support at $b=0$ with current reports, staff rollout is 0.053 (0.007 to 0.099) better than human transport (descriptive). When workers are mostly available, the forecast based dispatch thus uses robots without a measurable service cost, but it admits very few missions.
	
	\subsection{Volume and selection}\label{sec:vs}
\begin{figure}[!htbp]
		\centering
		\includegraphics[width=0.85\textwidth]{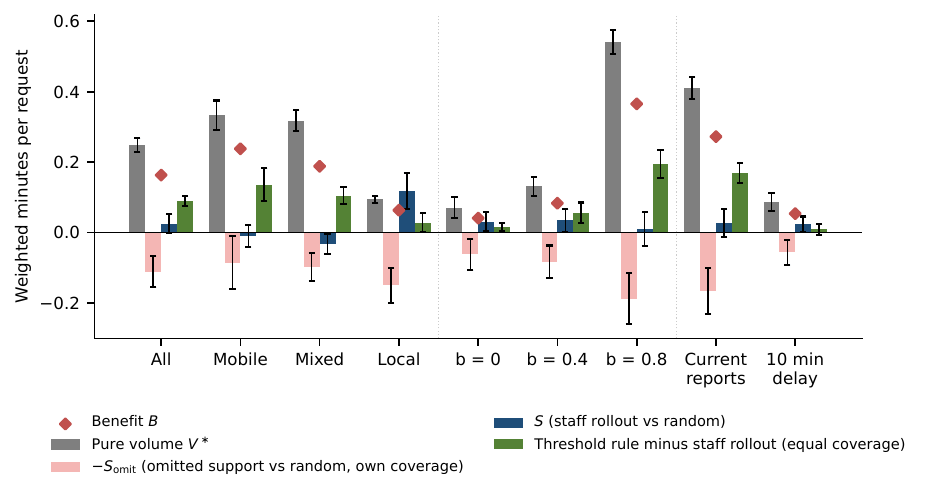}
		\caption{Benefit of representing support, $B$ (diamonds), its decomposition against the random admission curve into the pure volume effect $V^\ast$, the selection of omitted support at its own coverage (shown as $-S_{\mathrm{omit}}$), and the selection of staff rollout $S$, and the tardiness of the threshold rule minus staff rollout at equal coverage (green), with 95\% intervals, by subset; 20 seeds.}\label{fig:decomp}
	\end{figure}

	Matched random reaches 0.77\% coverage against 0.74\% for staff rollout (median ratio 1.03, 0.77 to 2.04), and the random arm at the coverage of omitted support reaches 2.09\% against 1.89\% (median ratio 0.90, 0.83 to 1.27). With a single coverage slope, Eq.~\eqref{eq:decomp} splits $B=0.163$ into $V=0.123$ (0.088 to 0.158) and $S=0.040$ (0.014 to 0.067), and Eq.~\eqref{eq:three} gives $V^\ast=0.286$ and $S_{\mathrm{omit}}=0.163$; these values are sensitive to the coverage errors, which are largest where $J$ rises most steeply with coverage. Table~\ref{tab:decomp} and Figure~\ref{fig:decomp} therefore report the decomposition against the random admission curve of each condition. The pure volume effect is $V^\ast=0.248$ (0.228 to 0.269), larger than the total benefit. Omitted support selects better than random admission at its own coverage, $S_{\mathrm{omit}}=0.111$ (0.066 to 0.155), which offsets part of the volume effect, and the selection value of staff rollout against random admission falls to $S=0.026$ ($-0.001$ to 0.052), about 9 weighted minutes per day. Against random admission, the benefit of representing support is therefore mainly a volume effect.
	
	Random admission is, however, not the relevant alternative to staff rollout; the forecast without support is. The comparison at equal coverage separates the two. The threshold rule reaches 0.73\% coverage against 0.74\% for staff rollout, and its tardiness is $J=1.170$, which is 0.090 (0.076 to 0.104) worse than staff rollout and 0.062 (0.027 to 0.097) worse than random admission at the same coverage. Along the path of Eq.~\eqref{eq:thr}, $B$ thus splits into 0.074 (0.056 to 0.091) from trimming omitted support to the coverage of staff rollout and 0.090 from selection at equal coverage, about 55\% of $B$. Omitted support appears better than random at its own coverage because most of its admissions carry no tardiness consequence: in the threshold rule runs, 91\% of the decisions in which the forecast without support prefers a robot are ties in weighted tardiness, and excluding these ties reduced its coverage from about 1.5\% to about 0.25\% in the pilot runs. Its surplus missions are thus mostly harmless, but when it must choose among the few missions that matter, it chooses worse than random, whereas staff rollout chooses better. Representing support therefore improves the choice of missions, not only their number.
	
	\begin{table}[!htbp]
		\caption{Staff rollout against human transport (SR $-$ H), the three-term decomposition $B=V^\ast-S_{\mathrm{omit}}+S$ (Eq.~\ref{eq:three}) against the random admission curve of each condition, and the threshold rule minus staff rollout at equal coverage (Thr $-$ SR, Eq.~\ref{eq:thr}), by subset; weighted minutes per request, 20 seeds; means over 20 seeds with 95\% confidence intervals (Student $t$ across seeds) in brackets below. All intervals of $V^\ast$ exclude zero; those of SR $-$ H are given in Section~\ref{sec:results}. Positive SR $-$ H favors human transport; positive $S$ and $S_{\mathrm{omit}}$ mean that the forecast selects better than random admission at its own coverage; positive Thr $-$ SR means that staff rollout selects better than the forecast without support at equal coverage.}\label{tab:decomp}
		\centering
		\setlength{\tabcolsep}{3pt}
		\begin{tabular}{p{1.6cm}llp{2.8cm}lp{2.45cm}p{2.45cm}}
			\toprule
			Subset & SR $-$ H & $B$ & $S$ & $V^\ast$ & $S_{\mathrm{omit}}$ & Thr $-$ SR \\
			\midrule
			All & 0.113 & 0.163 & 0.026 $[-0.001,0.052]$ & 0.248 & 0.111 $[0.066,0.155]$& 0.090 $[0.076,0.104]$\\
			Mobile & 0.083 & 0.238 & $-0.010$ $[-0.040,0.021]$& 0.333 & 0.085 $[0.011,0.159]$& 0.136 $[0.090,0.182]$\\
			Mixed & 0.099 & 0.188 & $-0.032$ $[-0.062,-0.002]$& 0.317 & 0.097 $[0.058,0.136]$& 0.105 $[0.080,0.129]$\\
			Local & 0.158 & 0.064 & 0.119 $[0.067,0.170]$& 0.095 & 0.150 $[0.100,0.200]$& 0.028 $[0.001,0.055]$\\
			$b=0$ & $-0.003$ & 0.041 & 0.032 $[0.004,0.059]$& 0.071 & 0.062 $[0.017,0.106]$& 0.017 $[0.005,0.029]$\\
			$b=0.4$ & 0.010 & 0.084 & 0.035 $[0.003,0.067]$& 0.132 & 0.083 $[0.036,0.130]$& 0.056 $[0.027,0.086]$\\
			$b=0.8$ & 0.332 & 0.366 & 0.011 $[-0.037,0.058]$& 0.542 & 0.187 $[0.116,0.259]$& 0.196 $[0.155,0.236]$\\
			Current reports & 0.209 & 0.273 & 0.027 $[-0.012,0.067]$& 0.411 & 0.165 $[0.099,0.231]$& 0.170 $[0.142,0.197]$\\
			Ten-minute delay &0.018 & 0.054 & 0.024 $[0.002,0.046]$& 0.086 & 0.056 $[0.021,0.091]$& 0.010 $[-0.006,0.025]$\\
			\bottomrule
		\end{tabular}
	\end{table}
	
\begin{figure}[!htbp]
		\centering
		\includegraphics[width=\textwidth]{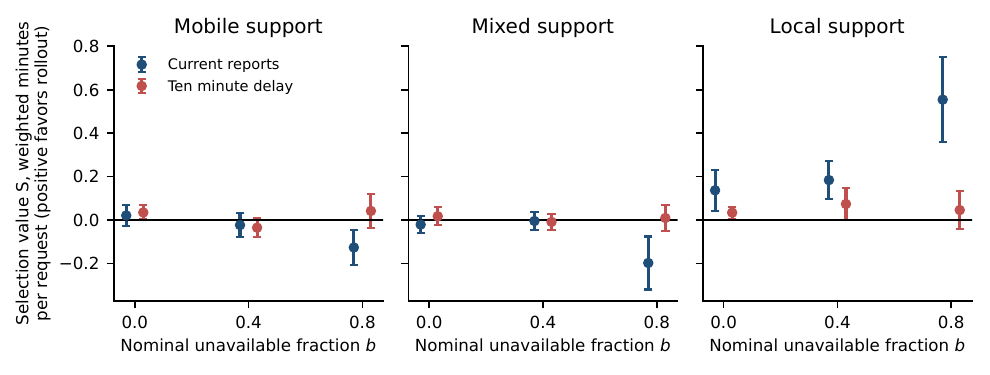}
		\caption{Selection value against matched random, $S=J_{\mathrm{rand}}-J_{\mathrm{staff}}$, without coverage adjustment in the 18 conditions (means over 20 seeds with individual 95\% intervals). Positive values favor staff rollout.}\label{fig:selection}
	\end{figure}

	The two paths agree on the volume effect but differ on selection (Table~\ref{tab:decomp}; Fig.~\ref{fig:decomp}). With mobile and mixed support, staff rollout selects no better than random admission ($S=-0.010$ and $-0.032$) but far better than the threshold rule at equal coverage ($J_{\mathrm{thr}}-J_{\mathrm{staff}}=0.136$ and $0.105$). With local support, both forecasts select better than random admission ($S=0.119$, $S_{\mathrm{omit}}=0.150$) and differ little at equal coverage (0.028, 0.001 to 0.055), because omitting support costs little when workers are stationary. As unavailability rises, $B$ grows from 0.041 to 0.366, mostly through the volume effect $V^\ast$, and the gap between the two forecasts at equal coverage widens from 0.017 at $b=0$ to 0.196 at $b=0.8$; with ten-minute reports it disappears (0.010, $-0.006$ to 0.025). Figure~\ref{fig:selection} shows the selection value against random admission in each condition. After an exploratory Holm correction over the 18 tests, staff rollout selects better than random in two local conditions and worse in the mobile and mixed conditions at $b=0.8$ with current reports. Without the coverage adjustment, the local mean (0.171) is dominated by one condition, $b=0.8$ with current reports, in which $S=0.554$; in the other five local conditions $S$ ranges from 0.03 to 0.18. Against the random admission curve, the same condition falls to 0.218 and the local mean to the 0.119 of Table~\ref{tab:decomp}, because matched random there exceeded the coverage of staff rollout where $J$ rises steeply with coverage.

	\subsection{Which missions do the forecasts select?}\label{sec:missions}
\begin{figure}[!htbp]
		\centering
		\includegraphics[width=\textwidth]{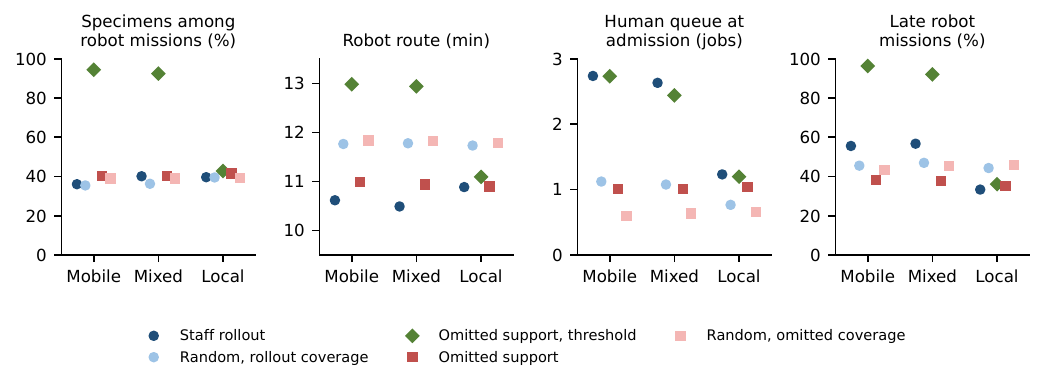}
		\caption{Robot missions selected by staff rollout, omitted support, the threshold rule (omitted support at the coverage of staff rollout), and matched random at the coverage of each forecast, by support arrangement: share of specimens, deterministic robot route duration, length of the human queue at the admission decision, and share of late robot missions. Pooled over 20 seeds and the six conditions of each arrangement.}\label{fig:profile}
	\end{figure}

	Figure~\ref{fig:profile} profiles the robot missions chosen by the five admission rules (values in Table~A5 of the online appendix). Neither forecast keeps urgent requests away from robots: specimens make up 39\% of the robot missions of staff rollout and 38\% of those of matched random at the same coverage, a difference of 0.8 percentage points ($-3.0$ to 4.6), and the shares under omitted support and its random arm are similar (Table~A5). Both forecasts do, however, choose shorter missions than random admission, by 0.94 minutes (0.72 to 1.17) of deterministic route time for staff rollout and by 0.87 minutes (0.69 to 1.05), or about 50~m of robot travel, for omitted support.
	
	The rules differ in when they admit robots. Staff rollout admits a robot when the human queue holds 0.72 (0.53 to 0.91) more jobs than at a random admission, and 1.39 (0.96 to 1.81) and 1.29 (0.94 to 1.64) more jobs with mobile and mixed support, when only 2.5 and 2.3 mobile workers are idle instead of 6.5 and 6.6 under random admission. These differences are paired means over seeds; the pooled values of Table~A5 of the online appendix weight each seed by its number of robot missions and therefore differ slightly (1.62 and 1.55). Because such robots need the same busy transporters for their handovers, 56\% of their missions are late with mobile support and 57\% with mixed support, against 46\% and 47\% under random admission; these late shares exceed by an order of magnitude the 5\% of all requests that are late, and their tardiness offsets the advantage of the shorter routes. The rule admits a robot when the robot lowers the predicted tardiness of the whole queue (Eq.~\ref{eq:score}), which happens mostly when the queue is long, and precisely then the mission must wait for its own support. Local support does not depend on transporters for handovers, so the timing effect is smaller (0.44 jobs) and only 33\% of the robot missions of staff rollout are late, against 44\% under random admission. Omitted support, which sees no support cost, admits robots closer to the average state (0.38 jobs above random admission), and its robot missions are late less often than under random admission in every arrangement.
	
	Splitting tardiness into the part incurred by robot-served requests (own) and by human-served requests (other) locates the selection value. Of $S=0.040$, the own part is 0.033 (0.012 to 0.054) and the other part 0.008 ($-0.004$ to 0.020); with local support, the own part is 0.130 and the other part 0.041. Of $S_{\mathrm{omit}}=0.163$, the own part is 0.143 (0.113 to 0.173). The pure volume effect, in contrast, also delays other work when transporters provide the support: of the unadjusted $V^\ast=0.387$ with mobile support, 0.085 (0.048 to 0.122) falls on human-served requests, whereas with local support the other part is slightly negative ($-0.008$).

	The threshold rule shows what goes wrong without a support forecast. With mobile and mixed support, 95\% and 93\% of its robot missions carry specimens on routes of about 13 minutes, admitted when the human queue is as long as under staff rollout (2.7 and 2.4 jobs); 96\% and 92\% of these missions are late, with a mean weighted tardiness of 70 and 62 minutes per robot mission, against 24 and 27 minutes under staff rollout. A forecast that ignores support expects the largest gain from giving an urgent request to a robot exactly when the transporters are busy, but the robot then waits for the same transporters. Staff rollout foresees this cost and admits other missions at those times. With local support, the threshold rule behaves much like staff rollout, with 43\% specimens and 36\% late missions. Of the difference of 0.090 at equal coverage, 0.098 (0.085 to 0.112) arises in the robot missions themselves and $-0.009$ ($-0.017$ to 0.000) in human-served requests.

	\subsection{Reporting delay}
	The same mechanism explains the effect of reporting delay. With ten-minute reports, the prediction error of staff rollout grows from 2.20 to 3.39 minutes, yet its weighted tardiness falls by 0.191 (0.159 to 0.222), because its coverage drops from 1.15\% to 0.33\%. Matched random at these coverages falls by 0.225, and the change in $S$ between the two delays does not differ from zero ($-0.035$, $-0.075$ to 0.006). With delayed reports, the two forecasts select equally well at equal coverage (Table~\ref{tab:decomp}): the support model improves selection only while the worker state is current, and older information helps only by admitting fewer robots. Dispatch rules therefore cannot be ranked by their prediction accuracy alone.
	
	\subsection{Staff time and service}
\begin{figure}[!htbp]
		\centering
		\includegraphics[width=\textwidth]{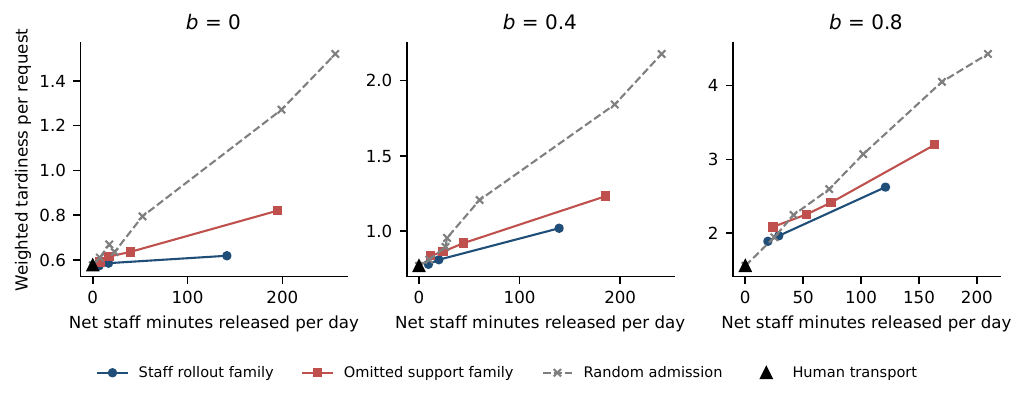}
		\caption{Weighted tardiness against net released staff minutes per day by unavailable fraction $b$. The staff rollout family joins the unrestricted rule and the allowance rules with $\varphi=3$ and 10; the omitted support family adds the threshold rule; random admission joins human transport (probability zero) and the six random arms. Means over 20 seeds and six conditions per panel.}\label{fig:front}
	\end{figure}

	Since robots are introduced to relieve staff, weighted tardiness must be read against the staff time that a policy releases (Fig.~\ref{fig:front}; values in Table~A1 of the online appendix). Unrestricted staff rollout releases 11.9 staff minutes per day at a mean cost of 0.113 in weighted tardiness relative to human transport. At $b=0$, a few minutes per day are released without a detectable increase in tardiness, whereas at $b=0.8$ every robot policy is clearly worse than human transport, and tardiness rises as more time is released. With $\varphi=10$, the support-aware rule serves 8.7\% of requests and releases 134 minutes per day (about 2.2 hours) at a mean cost of 0.453. Random admission at a similar coverage (9.7\%) costs far more ($J=2.39$), so at high coverage both forecasts select much better than random: the curve-adjusted selection value is 0.817 (0.768 to 0.867) for the support-aware rule with $\varphi=10$ and 0.897 (0.828 to 0.966) for the rule with omitted support at its coverage of 11.7\%.
	
	The two families also differ from each other. Interpolated along the family with omitted support to the same coverage, the support-aware rules have lower weighted tardiness by 0.085 (0.062 to 0.107) at the coverage of $\varphi=3$ (1.5\%) and by 0.169 (0.146 to 0.192) at that of $\varphi=10$ (8.7\%). The difference holds at every level of unavailability and with mobile and mixed support, but not clearly with local support (0.033, $-0.002$ to 0.067). Interpolated on each seed to the same released time, the support-aware family is better by 0.125 (0.107 to 0.143) at 60, 0.151 (0.134 to 0.167) at 100, and 0.163 (0.144 to 0.182) at 120 staff minutes per day. Across the whole range examined, from the 0.74\% coverage of unrestricted staff rollout to the 8.7\% of the allowance rule with $\varphi=10$, representing support thus improves the selection of robot missions, and the improvement grows with the number of missions.
	
	\subsection{Exact scheduling at matched volume}
	The optimizer admits fewer robots than staff rollout (0.52\% against 0.74\% of requests), and Table~\ref{tab:opt} decomposes their difference by Eq.~\eqref{eq:opt}. With two robots, the volume term $V_o=0.050$ accounts for all of the total difference of 0.045, and the selection value of the optimizer, $S_o=0.036$, is slightly below that of staff rollout, $S_r=0.040$. With four and eight robots, the total difference grows to 0.081 and 0.146, but it remains a volume effect: $V_o$ rises to 0.139 and 0.279, neither selection value differs from zero, and with eight robots random admission at the coverage of the optimizer even beats the optimizer. Exact scheduling of the known support thus adds no detectable selection value beyond that of the rollout. Unlike the forecast with omitted support (Section~\ref{sec:vs}), the optimizer was not run as a threshold version, so its selection cannot be compared directly with that of staff rollout at equal coverage. An admission decision takes 205~ms with two robots and 514~ms and 1.2~s with four and eight robots, against about 1~ms for staff rollout. The solver reached the optimum of the primary objective in about 98\% of the cases, and fewer than 0.001\% of the cases fell back on the heuristic.

	\begin{table}[!htbp]
		\caption{Optimizer against staff rollout at matched volume. $R_o$, $R_r$: matched random at the coverage of the optimizer and of staff rollout. Two robots: 20 seeds, 18 conditions, 4 and 8 robots: 6 seeds, scenario conditions. Last three columns with 95\% intervals.}\label{tab:opt}
		\centering\footnotesize
		\setlength{\tabcolsep}{3pt}
		\begin{tabular}{lrrrrrrrr}
			\toprule
			Robots & Opt. & $R_o$ & Rollout & $R_r$ & Cov. opt./roll. (\%) & $V_o$ & $S_o$ & $S_r$ \\
			\midrule
			Two & 1.036 & 1.071 & 1.081 & 1.121 & 0.52 / 0.74 & 0.050 & 0.036 & 0.040 \\
			& & & & & & $[0.038,0.062]$ & $[0.001,0.070]$ & $[0.014,0.067]$ \\
			Four & 1.682 & 1.724 & 1.763 & 1.863 & 1.37 / 1.94 & 0.139 & 0.042 & 0.100 \\
			& & & & & & $[0.046,0.232]$ & $[-0.214,0.298]$ & $[-0.082,0.281]$ \\
			Eight & 1.942 & 1.805 & 2.088 & 2.083 & 2.10 / 3.09 & 0.279 & $-0.137$ & $-0.005$ \\
			& & & & & & $[0.143,0.414]$ & $[-0.436,0.161]$ & $[-0.207,0.197]$ \\
			\bottomrule
		\end{tabular}
	\end{table}
	
	\subsection{Class mix}\label{sec:mix}
	With 20\% instead of 40\% urgent specimens, weighted tardiness falls by about half under every policy (human transport 0.529 against 1.059 on the same ten seeds), but staff rollout admits robots for 0.75\% of requests, as with 40\% specimens (0.74\%), so the robots do not take over more of the routine transports (Table~A4 of the online appendix). The gap to human transport narrows in absolute terms, from 0.117 to 0.075 (difference $-0.042$, $-0.075$ to $-0.010$), but grows slightly as a share of the tardiness of human transport (14\% against 11\%). The benefit of representing support falls from 0.165 to 0.096 (difference $-0.068$, $-0.091$ to $-0.045$), whereas the comparison at equal coverage does not change: the threshold rule is 0.099 (0.077 to 0.121) worse than staff rollout, against 0.089 (0.069 to 0.109) with 40\% specimens on the same seeds (difference 0.010, $-0.008$ to 0.028), and 0.077 (0.037 to 0.117) worse than matched random. Trimming omitted support to the coverage of staff rollout no longer helps ($-0.002$, $-0.023$ to 0.019), so with fewer urgent requests the whole benefit of representing support is selection at equal coverage. The mechanism is the same as before. With mobile and mixed support, the threshold rule gives robots specimens in 68\% of its missions although specimens are only 20\% of requests, and 85\% of these missions are late, while staff rollout gives specimens 20\% of its missions and matched random 17\%; 48\% of the missions of staff rollout are admitted when no transporter is idle, against 20\% under matched random. Against random admission, the selection value of staff rollout is $S=0.022$ ($-0.010$ to 0.054, unadjusted), against 0.039 with 40\% specimens on the same seeds, a difference that cannot be distinguished from zero. Matched random reached a median of 95\% of the target coverage (85 to 272\%) and the threshold rule 101\% (36 to 205\%). The results on selection and on the timing of calls therefore do not depend on the assumed share of urgent requests, whereas the size of the volume effect does.

	\subsection{What the rules ask of staff}\label{sec:staff}
	Table~\ref{tab:staff} reports the staff-side measures of Section~\ref{sec:design}. Staff rollout generates 6.5 calls per day for its two robots, 2.5 per mission, against 16.5 under omitted support and about 100 under the idle pair rule and the allowance rule with omitted support at $\varphi=10$. A call of staff rollout waits 6.3 minutes until a worker arrives, longer than under matched random (4.9 minutes) and omitted support (5.0 minutes), and with mobile support 8.1 minutes against 5.8 under matched random; summed over the day, the robots of staff rollout wait 36 minutes for people, as long as those of matched random, which make slightly more but shorter calls.

	The longer waits follow from the timing found in Section~\ref{sec:missions}. With mobile and mixed support, 46\% of the robot missions of staff rollout are admitted when no transporter is idle and 32\% when three or more jobs wait in the human queue, against 18\% and 11\% under matched random at the same coverage; the difference in the first share is 26 percentage points (21 to 31). The threshold rule raises the share admitted without an idle transporter by a further 14 points (9 to 19), to 59\%, and omitted support at its own coverage admits 27\% of its missions in this state, 16 points (15 to 18) more than its random arm. A rule that minimizes tardiness therefore addresses most of its requests for help to transporters who are all occupied. It also leaves the robots almost unused at night, when fewer transporters are on duty: 1.2\% of the missions of staff rollout start in the night shift, against 6.0\% under matched random and 13.5\% under the threshold rule.

	Released time overstates the relief that robots bring to transporters. With mobile support, staff rollout releases 4.7 staff minutes per day, yet the other duties displaced by delivery and support work are the same as under human transport (difference 0.0, $-6.3$ to 6.3 minutes per day), and omitted support releases 30.9 minutes per day while it displaces 11.2 (5.8 to 16.6) minutes per day more of the transporters' other duties than human transport; the threshold rule displaces 4.8 ($-0.7$ to 10.4) more. Released time counts minutes, whereas displacement depends on when the work falls, and the support jobs of these rules, with their approach walks and waits, occupy transporters when their other duties are due. Only where local workers support the robots are the transporters relieved: under staff rollout their displaced duties fall by 55 minutes per day, while those of the 95 local workers rise by 2.3 minutes per day in total, or by 11.4 with the allowance rule at $\varphi=10$. Overtime past the end of shifts changes by less than one minute per day under every forecast rule.

	The availability gate of Section~\ref{sec:gate} tests whether a dispatcher can keep its calls away from busy staff. It halves the calls of staff rollout to 3.6 per day and the time robots wait for people to 15.5 minutes per day, and with mobile and mixed support it lowers the share of missions admitted without an idle transporter from 46\% to 21\%, close to the 18\% of random admission; the remaining share arises because the gate sees the reported, possibly delayed state. Service improves at the same time. The gate reaches $J=0.994$, which is 0.086 (0.072 to 0.101) below staff rollout with all 20 seeds favorable, below the optimizer (1.036), and only 0.027 (0.011 to 0.042) above human transport; with mixed support it does not differ from human transport (0.003, $-0.016$ to 0.023). The gain arises at $b=0.8$, where the gate is better than staff rollout by 0.254 (0.214 to 0.294); at $b=0$ and $b=0.4$ the two rules differ by less than 0.005, and with ten-minute reports the gate rarely binds. Against the random admission curve, formed in each condition by human transport, matched random, and the random arm at the coverage of omitted support, most of the gain is a volume effect: the gate admits robots for 0.40\% of requests instead of 0.74\%, and the curve accounts for 0.086 (0.063 to 0.109) of its improvement. With mobile and mixed support, however, the gate also selects better than staff rollout, by 0.027 (0.007 to 0.047) and 0.044 (0.018 to 0.071) relative to random admission, whereas with local support it blocks useful missions and selects worse ($-0.071$, $-0.138$ to $-0.003$). Its price is released time, which falls from 11.9 to 6.5 minutes per day. Under mobile support the gate reduces the transporters' displaced duties below those of human transport, by 7.4 (2.1 to 12.7) minutes per day, which among the forecast rules only the optimizer also achieves.

	\begin{table}[!htbp]
		\caption{What the rules ask of staff; main experiment, 20 seeds and 18 conditions, means over equally weighted conditions. Calls: loading, receiving, and cleaning assignments of workers to robots. Call to arrival: minutes from the robot's need for a worker to the start of service, including the wait for a free worker and the approach walk. Robot wait: call-to-arrival time summed over a day, minutes. No idle tr.: share of robot missions admitted while no transporter was idle, mobile and mixed support pooled (mission logs; --, not logged). Displaced: other duties of transporters displaced by delivery and support work, difference to human transport under mobile support. Released: net released staff minutes per day.}\label{tab:staff}
		\centering\footnotesize
		\setlength{\tabcolsep}{2.2pt}
		\begin{tabular}{lrrrrrrr@{\hspace{8pt}}r}
			\toprule
			Policy & \makecell[r]{Missions\\per day} & \makecell[r]{Calls\\per day} & \makecell[r]{Call to\\arrival} & \makecell[r]{Robot\\wait} & \makecell[r]{No idle\\tr. (\%)} & Released & Displaced & $J$ \\
			\midrule
			Human transport & 0 & 0 & -- & 0 & -- & 0 & 0 & 0.968 \\
			Staff rollout & 2.6 & 6.5 & 6.3 & 36.1 & 46 & 11.9 & 0.0 & 1.081 \\
			Availability gate & 1.4 & 3.6 & 6.2 & 15.5 & 21 & 6.5 & $-7.4$ & 0.994 \\
			Own completion & 2.3 & 5.8 & 6.7 & 31.3 & -- & 11.0 & 0.8 & 1.069 \\
			Optimizer & 1.8 & 4.5 & 7.9 & 26.9 & -- & 8.4 & $-24.9$ & 1.036 \\
			Matched random & 2.7 & 6.7 & 4.9 & 36.2 & 18 & 14.2 & $-1.6$ & 1.121 \\
			Threshold rule & 2.6 & 6.8 & 6.3 & 39.5 & 59 & 14.3 & 4.8 & 1.170 \\
			Omitted support & 6.6 & 16.5 & 5.0 & 96.9 & 27 & 31.2 & 11.2 & 1.244 \\
			Support aware, $\varphi=10$ & 30.1 & 71.9 & 3.7 & 244.2 & -- & 133.9 & $-3.7$ & 1.421 \\
			Omitted support, $\varphi=10$ & 40.3 & 98.9 & 4.0 & 366.7 & -- & 181.3 & $-3.0$ & 1.750 \\
			Idle pair rule & 40.8 & 100.0 & 4.0 & 339.0 & -- & 204.7 & $-14.0$ & 1.804 \\
			\bottomrule
		\end{tabular}
	\end{table}

	\subsection{When could robots help?}
\begin{figure}[!htbp]
		\centering
		\includegraphics[width=0.55\textwidth]{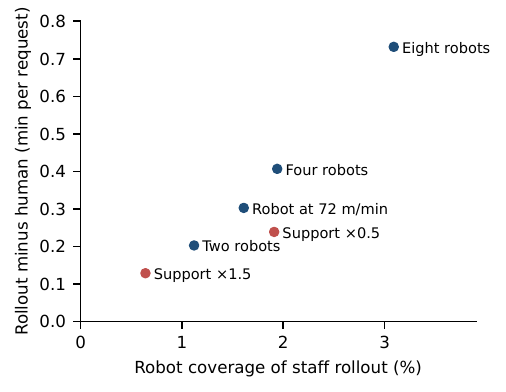}
		\caption{Staff rollout minus human transport against the robot coverage of staff rollout in the sensitivity runs; 6 seeds, scenario conditions. Red points change the support time per mission.}\label{fig:gap}
	\end{figure}

	Figure~\ref{fig:gap} relates the gap between staff rollout and human transport to robot coverage in the sensitivity runs (Table~A2 of the online appendix). More robots, and even a robot as fast as a person, widen the gap because they raise coverage; longer support times narrow it because they lower coverage. Halving the support time, in contrast, raises coverage with a smaller gap per mission, which points to the staff time per mission as the lever.
	
\begin{figure}[!htbp]
		\centering
		\includegraphics[width=0.85\textwidth]{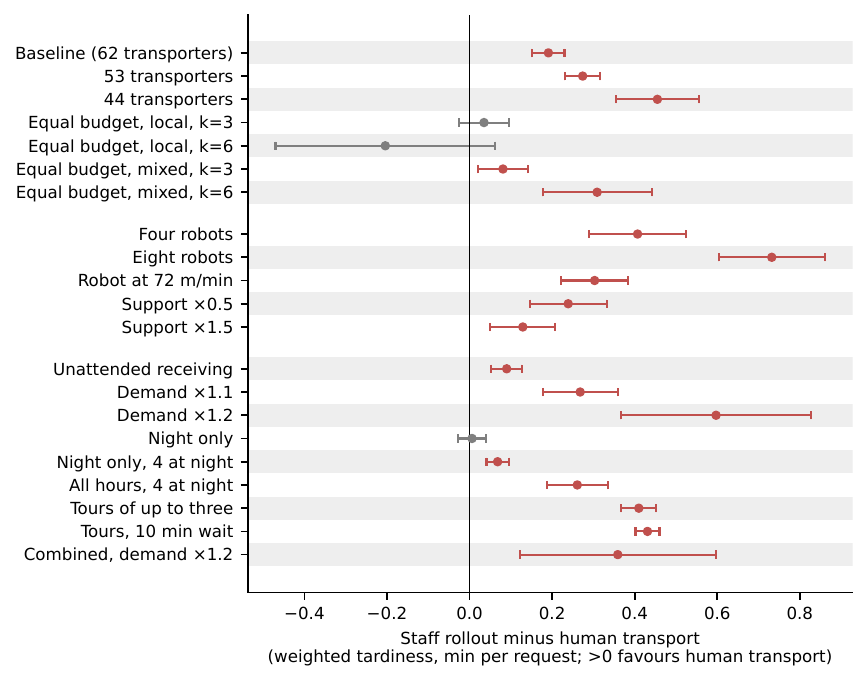}
		\caption{Staff minus human transport with 95\% confidence intervals. Positive values indicate that human transport is better. Sensitivity has 6 seeds, and other scenarios have 12 seeds. The rows with equal budgets compare the results with human transport with the reduced workforce. Grey indicates that the confidence interval contains 0. All comparisons are descriptive.}\label{fig:forest}
	\end{figure}

	Figure~\ref{fig:forest} reports the scenarios, and Table~A3 of the online appendix reports the coverage, run time, and selection value. In the scarcity and unattended receiving scenarios, matched random reaches 8 to 14\% more coverage than staff rollout, and in the demand, tour, and combined scenarios 10 to 40\% more, which biases $S$ upwards. Because the scenarios have no random admission curve, we report $S_{\mathrm{adj}}$ with a mean slope of $\beta=0.215$ (0.163 to 0.261) weighted minutes per request per percentage point of coverage, estimated from the arms $R_o$ and $R_r$ of Eq.~\eqref{eq:opt}. This slope was estimated at baseline demand over all levels of unavailability, whereas the scenarios use $b\in\{0.4,0.8\}$, where the random admission curve is steeper, so the adjustment is probably too small.
	
	\emph{Scarcity leaves no role for robots.} Reducing the roster to 53 and 44 transporters raises the tardiness of human transport to 1.340 and 2.269, and the gap between staff rollout and human transport grows from 0.191 to 0.274 and 0.455, because the robots draw on the same scarce transporters; with 44 transporters, mobile support, and $b=0.8$ it reaches 1.604. When the staffed hours are held at 496 by replacing transporters with three or six local workers, staff rollout reaches the level of human transport on the reduced roster only with six local workers ($-0.204$, $-0.470$ to 0.062), where the selection value is large and hardly changed by the coverage adjustment ($S_{\mathrm{adj}}=0.390$). No equal budget variant reaches human transport with the full workforce.
	
	\emph{Unattended receiving} halves the gap to 0.090 and releases more staff time (24.0 against 18.2 minutes per day). The volume penalty disappears ($V=-0.014$, $-0.050$ to 0.023): once a mission takes little staff time, admitting more robots no longer costs service.
	
	\emph{Higher demand} widens the mean gap to 0.268 and 0.597, yet with local support at $b=0.4$ staff rollout becomes better than human transport, by 0.184 (0.072 to 0.295) and 0.403 (0.062 to 0.744). The unadjusted selection value doubles, but after the coverage adjustment it rises only from 0.060 to 0.138 and 0.083.
	
	\emph{Night-only operation} closes the gap with the full roster (0.006, $-0.028$ to 0.040), but only because robots then serve 0.16\% of requests. With four night transporters, night-only robots are worse than human transport by 0.068 (0.041 to 0.096), less than robots operating at all hours (0.261), but they do not compensate for the missing night staff.
	
	\emph{Multi-load tours} carry 1.6 to 1.7 loads per tour but widen the gap to 0.410, and to 0.431 when the robot holds for up to 10 minutes. Staff rollout with tours is worse than without them by 0.219 (0.175 to 0.262), because loads that join a tour bypass the admission forecast while each of them still needs staff at both ends. The \emph{combined} scenario (demand $\times1.2$, unattended receiving, and tours with holding) releases 76 staff minutes per day, about four times the baseline, and narrows the gap at this demand from 0.597 to 0.359; with local support at $b=0.4$, however, staff rollout is worse than human transport by 0.827 (0.330 to 1.323). Here omitted support is slightly better than staff rollout (6.104 against 6.135), the only scenario of the study with a negative $B$, and most of the unadjusted selection value (0.315) vanishes after the coverage adjustment (0.089).
	
	\section{Discussion}\label{sec:discussion}
	\subsection{What a support forecast does}
	
	The answer to RQ1 depends on the reference. Against random admission, the pure volume effect exceeds the total benefit: the average robot mission is worse than a human delivery, so admitting fewer robots always helps, and omitted support even selects better than random at its own coverage. Against omitted support at the same coverage, in contrast, more than half of the benefit is selection. The reason lies in how omitted support gains its coverage. Most of its robot admissions are tardiness ties broken by flow time, low-cost admissions in calm periods that make the rule appear selective; among the few admissions that matter, those with the largest predicted tardiness gain, ignoring support produces the most mistakes, because it sends urgent requests to robots while the transporters needed to load and receive them are busy. For a dispatcher that minimizes tardiness, the value of representing support therefore lies in a small number of high-impact decisions, and along the staff time front this value grows as more missions are admitted.
	
	For evaluation, volume-matched random controls are thus necessary but not sufficient. They show what any rule with the same admission rate would achieve, but random admission is a weak competitor that a poorly ranking rule can beat through harmless low-impact admissions; whether volume explains the full effect can only be decided by comparing the competing rule at the same coverage.
	
	The mission logs show what selection consists of. It does not come from more accurate forecasts, since the prediction error of staff rollout is almost the same in the three arrangements (2.74, 2.81, and 2.83 minutes), nor from keeping urgent requests with people, since the class mix of robot missions equals that of random admission. Both forecasts prefer shorter routes, which shortens the slower robot leg. What separates the rules is timing. Because the support-aware rule admits a robot only when this lowers predicted tardiness, it admits robots when the human queue is long, and with mobile transporters this is exactly when the robot's own handovers must wait for the same busy people; against random admission, the advantage of shorter routes is then lost at the endpoints. The forecast without support, restricted to the same number of missions, makes this mistake far more often, because it does not see that the robot needs the busy transporters at all. With stationary local support the timing is harmless, robot missions are late less often than under random admission, and both forecasts select well. Almost all selection value lies in the robot missions' own tardiness: forecasting the effect of a mission on other work contributes little. When urgent specimens make up 20\% instead of 40\% of requests, the comparison at equal coverage is unchanged and the forecast without support still gives its decisive missions to specimens (Section~\ref{sec:mix}), so the mechanism does not hinge on the assumed class mix.

	The optimizer points the same way. Exact scheduling of the known work selects about as well as staff rollout when both are measured against random admission at their own coverage ($S_o=0.036$ against $S_r=0.040$ with two robots, neither different from zero with four and eight); its improvement over staff rollout is a volume effect and costs two orders of magnitude more computation per decision. Since deterministic schedules of the known work are already adequate, the limiting factor is the uncertain availability of staff rather than the scheduling of work that is known.
	
	\subsection{Staff time as the decisive parameter}
	Robots are introduced in hospitals to reduce the workload of staff, and every policy examined here that releases staff time does so at a cost in service. Rules that forecast support explicitly buy released time more cheaply than rules with omitted support, which matters most once robots serve more than a few missions per day. In the main experiment, staff rollout gives the two robots only 0.74\% of requests, about 2.6 missions per day or 1.3 per robot, and releases 11.9 staff minutes per day. From the perspective of hospital management, such a fleet would hardly justify its cost, and this result is the strongest support for the central claim of the paper: without infrastructure that removes the human handover, intelligent dispatch merely limits the losses that robots cause instead of making them useful. The scenarios point the same way. Robots become competitive with human transport only when the staff time per mission is reduced or staff are available where the robots arrive. Staff shortages, a fixed staff budget, higher demand, more or faster robots, night-only operation, and tours do not achieve this, because they neither reduce the support time per mission nor add missions that avoid the same support. Unattended receiving halves the gap to human transport and removes the volume penalty, and with stationary local staff, robots dispatched by an explicit support forecast match or outperform human transport when staff are mostly available or demand is high.
	
	\subsection{The worker's side of robot support}\label{sec:hri}
	In the model, a worker who supports a robot is an approach walk, a coordination time of $2e^{Z}$ minutes, and a service time, yet even this thin representation shows a side of the rules that tardiness and released time hide. A rule that minimizes tardiness admits a robot when the robot lowers the predicted tardiness of the queue, which happens when the queue is long, and so it addresses most of its requests for help to transporters who are all occupied (Section~\ref{sec:staff}). Studies of help-seeking robots would say that it asks at the worst moment and without regard to who is available \cite{rosenthal2012,kimoto2022}. Each such call reaches a worker as one more request competing with other work, and where that worker is a nurse it is an interruption of the kind that clinical studies associate with errors \cite{westbrook2010}; frequent calls also add to the alerts that staff must filter \cite{cvach2012}. The cost of an interruption depends on what it interrupts \cite{mcfarlane2002}, and none of it enters the tardiness objective. Field studies report the same tension: the same delivery robot was welcomed or resented depending on the workload of the unit and the interruptions it caused \cite{mutlu2008}, staff renegotiated whether it was a machine or a colleague \cite{ljungblad2012}, and on a nuclear medicine ward the number of walks of nurses was associated with perceived interruption and stress \cite{warmbein2025}. Shared elevators add spatial conflicts between robots, transporters, patients, and beds that the queue model reduces to waiting time.

	Two results bear directly on how hospital robots should ask for help. First, released time is a poor proxy for the relief that staff experience. With mobile support, no forecast rule except the optimizer and the availability gate displaced detectably fewer of the transporters' other duties than human transport, and omitted support displaced more, although every rule released staff minutes in the account of the system. A hospital that evaluates robots by released time alone may report relief that its staff do not feel, which would be one reason for the mixed acceptance reported from long deployments \cite{hebesberger2017}. Second, respecting the availability of staff need not cost service. The availability gate, a rule that a human coordinator could state in one sentence, halved the calls to staff and the waiting of robots for people, brought the timing of calls close to that of random admission, and lowered tardiness below that of staff rollout and the optimizer. It did so mostly by admitting fewer robots and it released less staff time, so it is not a free improvement, but it shows that the balance between service, released time, and the interruption of staff can be set deliberately by the dispatcher instead of being left to the tardiness objective. Both results held when urgent requests were halved, since the timing of calls did not change with the class mix (Section~\ref{sec:mix}).

	These findings suggest how a support forecast can serve the people around the robots. A dispatcher that predicts when and where support will be needed can announce a call before the robot arrives, bound the number of calls per worker and hour, and prefer workers who are idle, much as a help-seeking robot chooses whom to ask \cite{rosenthal2012}; the wording of the request then matters as well \cite{srinivasan2016}. Predictable requests support appropriate trust in the robots \cite{lee2004} and leave staff the involvement and responsibility that caregivers expect \cite{klebbe2025}, and robot use must still fit work processes and the condition of patients \cite{brunett2025,ohneberg2025}. Allowing staff to postpone or decline a call, which the model does not represent, would give them the autonomy that the present rules deny. Field trials of such rules should report fluency measures, in particular the time robots wait for people and the delay they cause to human work \cite{hoffman2019}, together with perceived workload, interruption, and acceptance \cite{turja2019}; the simulation identifies which quantities matter, but it cannot replace studies with staff.

	\subsection{Implications for robotic system design}
	The results suggest an order of priorities for the design of hospital robot fleets. The largest lever is the human handover itself: unattended receiving, for example lockable drawers or receptacles that are emptied later, deserves investment before dispatch intelligence or fleet size, because staffed handovers are the bottleneck of the robot system \cite{ortenzi2021}. Where handovers remain staffed, the dispatcher should optimize an explicit trade-off between service and released staff time, for example through the allowance of Section~\ref{sec:policies}, since a tardiness objective alone admits robots for only a few missions per day. A rule that dispatches robots while staff are busy must, in addition, check that the robots do not need the same staff for their handovers, or it will admit robots at the worst time for their own support; the availability gate of Section~\ref{sec:gate} is the simplest such check, and in this model it improved service while it halved the calls to staff. Staff-side measures, such as calls per day, the time robots wait for people, and the state of the workforce when calls arrive, belong in the evaluation of a dispatcher next to tardiness and released time. Evaluation, finally, needs two kinds of control: random admission at equal volume, without which a policy that merely admits fewer robots may falsely appear to achieve superior mission selection, and the competing rules at equal coverage, without which a policy that ranks the critical missions poorly may appear as good as one that ranks them well.
	
	\subsection{Limitations}
	The study uses one hospital pathway with assumed task classes, deadlines, speeds, support durations, and staffing, and it has no hardware or field validation; the verification tests check internal consistency, not agreement with a real robot fleet. The model of human unavailability is its strongest simplification. Each worker follows a deterministic 30-minute cycle with a fixed unavailable fraction $b$, and the dispatcher expects a worker reported as unavailable back after a constant $15b$ minutes. Real wards produce bursts of urgent clinical work that withdraw several workers at once and for longer than one cycle, and staff decide for themselves whether to answer, postpone, or ignore a waiting robot. Bursts would make predicted return times less reliable and would concentrate robot waiting in the periods in which robots are least welcome, and the freedom of staff to postpone support adds variance that no deterministic continuation represents. We expect both features to raise the cost of staffed robot support and to make the timing of admissions matter more, but the present model cannot quantify either effect, and the displaced clinical work is not valued at all. The task classes are synthetic because the Freiburg data contain none: 40\% of requests are treated as urgent specimens with weight 4, whereas in many hospitals a large part of laboratory transport is scheduled. Halving this share left the coverage of staff rollout, the comparison at equal coverage, and the timing of calls unchanged and reduced the volume effect (Section~\ref{sec:mix}), but this check used one alternative mix and ten seeds and changed neither deadlines nor service times. The staff-side measures are quantities of the model: a call costs nothing beyond its minutes, every worker answers every call, and displaced duties are not weighted by their clinical importance, so the measures describe exposure to interruption, not workload or stress as staff perceive them. The availability gate and the staff-side analysis were designed after all other results, and the gate was compared with random admission on a curve of three points per condition instead of seven. Local workers add capacity, so arrangements are compared only through the equal budget variant. The random arms match coverage only approximately; the random admission curve corrects this by linear interpolation between seven points per condition, and the scenario phases rely on a single slope estimated at baseline demand. The threshold rule is one way of lowering the volume of omitted support while keeping its order of preference; a different ordering, for example by flow time alone, could select differently. The mission logs describe what the rules select, not the counterfactual outcome of each individual decision. The warm-up check compares fourteen with seven warm-up days in three conditions only and is not paired. All control, front volume, equal coverage, and scenario phases were designed after the main results; the scenario phases use 6 to 12 seeds and should be read as exploratory. Higher demand replicates observed requests within the same minute; unattended receiving ignores receptacle capacity and later collection; tours accept loads by a fixed rule the forecast does not anticipate; and the optimizer's wall-clock limits make its runs not exactly reproducible. Intervals reflect simulation noise only, not input uncertainty \cite{barton2014}.

	\section{Conclusion}\label{sec:conclusion}
	In a simulation of hospital robot dispatch on recorded requests, forecasting staff support reduced weighted tardiness relative to the same rollout with omitted support. Against random admission at matched coverage, this improvement appeared to be mainly a volume effect, and exact scheduling of the known work added no selection value beyond that of the rollout. Compared at equal coverage, the forecast with omitted support did worse, because the gains it predicted to be largest came from sending urgent requests to robots while the transporters who had to support them were busy, and along the staff time front the advantage of the support-aware rules grew with coverage. Representing human support therefore improves the choice of robot missions, most of all when mobile transporters support the robots and the staff state is reported without delay, and this held with half as many urgent requests. Seen from the staff side, the support-aware rule still addressed most of its calls to transporters who were all busy, and its released staff time did not reduce the transporters' displaced duties; a gate that admitted robots only when idle workers could serve them halved the calls to staff and lowered tardiness, mostly by admitting fewer robots. Human transport had the lowest mean weighted tardiness, but its advantage arose only when staff were mostly unavailable; elsewhere, forecast-based dispatch used robots without a detectable cost, and robots outperformed human transport only with local support, where staff were available or demand was high. Dispatch policies for hospital robots should be evaluated at equal admission volume, against random admission and against each other, and reported together with the staff time they release and with what they ask of the people who support the robots: how often they are called, how long robots wait for them, and when the calls arrive. The main design parameter is the staff time that each mission consumes, and the main open question is how staff experience the calls, which only field studies can answer.
	
		\section*{Acknowledgments}
We thank A.~Hess and D.~Stampa for publishing the Freiburg transport dataset used in this study. This work is partially financed from the research Grant No. 0211/SBAD/0226 by Poznan University of Technology.
	
	\section*{Declarations}
	
	\textbf{Funding.} This work was partially financed from the research Grant No. 0211/SBAD/0226 by Poznan University of Technology.
	
	\noindent
	\textbf{Competing interests.} The authors declare no competing interests.
	
	\noindent
	\textbf{Ethics approval and consent to participate.} Not applicable. The study uses public, de-identified transport records and synthetic task attributes; it involved no participants and no new collection of personal data.
	
	\noindent
	\textbf{Consent for publication.} Not applicable.
	
	\noindent
	\textbf{Data availability.} The request and pathway data are available at \url{https://doi.org/10.5281/zenodo.19085901}. The raw outputs of all 24,661 simulation runs (the 15,180 analysed runs of Table~\ref{tab:phases} and the warm-up check, pilot, calibration, verification, test, and probe runs described in Section~\ref{sec:design}), the frozen experiment designs, the calibration logs, and the summary analyses, including the staff-side measures of all phases, are available at \url{https://doi.org/10.5281/zenodo.24322157}.
	
	\noindent
	\textbf{Code availability.} The simulation code, tests, experiment runners, and analysis and figure scripts are available at \url{https://github.com/siwekk/hospital-robot-staff-support} under the MIT license. The repository also holds the online appendix (folder \texttt{appendix}) with the additional result tables (Appendix~A) and the STRESS-DES checklist (Appendix~B).
	
	\noindent
	\textbf{Author contributions.} Both authors contributed equally to the conception, modeling, analysis, and writing.
	
	\noindent
	\textbf{Use of generative AI.} 	During the preparation of this manuscript, the authors used large language model assistants solely for language copy-editing and minor code refactoring. The conceptual model, simulation logic, experimental design, and statistical analyses were developed by the authors, who thoroughly verified all code and text and take full responsibility for the content of the publication.
	
	\newpage
	\bibliographystyle{unsrtnat}
\bibliography{references}

@inproceedings{mutlu2008, author={Mutlu, B. and Forlizzi, J.}, title={Robots in organizations: The role of workflow, social, and environmental factors in human-robot interaction}, booktitle={Proceedings of the 3rd ACM/IEEE International Conference on Human-Robot Interaction}, pages={287--294}, year={2008}, doi={10.1145/1349822.1349860}}

@inproceedings{ozkil2009, author={{\"O}zkil, A. G. and Fan, Z. and Dawids, S. and Aan{\ae}s, H. and Kristensen, J. K. and Christensen, K. H.}, title={Service Robots for Hospitals: A Case Study of Transportation Tasks in a Hospital}, booktitle={Proceedings of the 2009 IEEE International Conference on Automation and Logistics}, pages={289--294}, year={2009}, doi={10.1109/ICAL.2009.5262912}}

@inproceedings{ljungblad2012, author={Ljungblad, S. and Kotrbova, J. and Jacobsson, M. and Cramer, H. and Niechwiadowicz, K.}, title={Hospital robot at work: Something alien or an intelligent colleague?}, booktitle={Proceedings of the ACM 2012 Conference on Computer Supported Cooperative Work}, pages={177--186}, year={2012}, doi={10.1145/2145204.2145233}}

@article{holland2021, author={Holland, J. and Kingston, L. and McCarthy, C. and Armstrong, E. and O'Dwyer, P. and Merz, F. and McConnell, M.}, title={Service robots in the healthcare sector}, journal={Robotics}, volume={10}, number={1}, pages={47}, year={2021}, doi={10.3390/robotics10010047}}

@article{kriegel2022, author={Kriegel, J. and Rissbacher, C. and Reckwitz, L. and Tuttle-Weidinger, L.}, title={The requirements and applications of autonomous mobile robotics ({AMR}) in hospitals from the perspective of nursing officers}, journal={International Journal of Healthcare Management}, volume={15}, number={3}, pages={204--210}, year={2022}, doi={10.1080/20479700.2020.1870353}}

@article{fragapane2023, author={Fragapane, G. and Hvolby, H.-H. and Sgarbossa, F. and Strandhagen, J. O.}, title={Autonomous mobile robots in sterile instrument logistics: an evaluation of the material handling system for a strategic fit framework}, journal={Production Planning and Control}, volume={34}, pages={53--67}, year={2023}, doi={10.1080/09537287.2021.1884914}}

@article{lopez2022, author={L{\'o}pez, J. and Zalama, E. and G{\'o}mez-Garc{\'i}a-Bermejo, J.}, title={A simulation and control framework for {AGV} based transport systems}, journal={Simulation Modelling Practice and Theory}, volume={116}, pages={102430}, year={2022}, doi={10.1016/j.simpat.2021.102430}}

@article{cheng2023, author={Cheng, L. and Zhao, N. and Yuan, M. and Wu, K.}, title={Stochastic scheduling of autonomous mobile robots at hospitals}, journal={PLOS ONE}, volume={18}, note={Article e0292002}, year={2023}, doi={10.1371/journal.pone.0292002}}

@inproceedings{vandonkelaar2026, author={Van Donkelaar, M. and Van Den Boom, T. and Martinez, C. and Boutin, O. and Bouzaiene, A. B.}, title={Modeling and Scheduling of Hospital Robot Fleets Using Switching-max-plus Algebra}, booktitle={Proceedings of the 18th International Workshop on Discrete Event Systems}, pages={277--282}, year={2026}, doi={10.1109/WODES69290.2026.11598284}}

@article{gombolay2018, author={Gombolay, M. and Yang, X. J. and Hayes, B. and Seo, N. and Liu, Z. and Wadhwania, S. and Yu, T. and Shah, N. and Golen, T. and Shah, J.}, title={Robotic assistance in the coordination of patient care}, journal={The International Journal of Robotics Research}, volume={37}, number={10}, pages={1300--1316}, year={2018}, doi={10.1177/0278364918778344}}

@article{gerkey2004, author={Gerkey, B. P. and Matari{\'c}, M. J.}, title={A formal analysis and taxonomy of task allocation in multi-robot systems}, journal={The International Journal of Robotics Research}, volume={23}, pages={939--954}, year={2004}, doi={10.1177/0278364904045564}}

@article{korsah2013, author={Korsah, G. A. and Stentz, A. and Dias, M. B.}, title={A comprehensive taxonomy for multi-robot task allocation}, journal={The International Journal of Robotics Research}, volume={32}, number={12}, pages={1495--1512}, year={2013}, doi={10.1177/0278364913496484}}

@article{nunes2017, author={Nunes, E. and Manner, M. D. and Mitiche, H. and Gini, M.}, title={A taxonomy for task allocation problems with temporal and ordering constraints}, journal={Robotics and Autonomous Systems}, volume={90}, pages={55--70}, year={2017}, doi={10.1016/j.robot.2016.10.008}}

@article{das2015, author={Das, G. P. and McGinnity, T. M. and Coleman, S. A. and Behera, L.}, title={A distributed task allocation algorithm for a multi-robot system in healthcare facilities}, journal={Journal of Intelligent and Robotic Systems}, volume={80}, pages={33--58}, year={2015}, doi={10.1007/s10846-014-0154-2}}

@article{galati2023, author={Galati, G. and Primatesta, S. and Rizzo, A.}, title={Auction-Based Task Allocation and Motion Planning for Multi-Robot Systems with Human Supervision}, journal={Journal of Intelligent and Robotic Systems}, volume={109}, pages={24}, year={2023}, doi={10.1007/s10846-023-01935-x}}

@article{ortenzi2021, author={Ortenzi, V. and Cosgun, A. and Pardi, T. and Chan, W. P. and Croft, E. and Kuli{\'c}, D.}, title={Object handovers: A review for robotics}, journal={IEEE Transactions on Robotics}, volume={37}, number={6}, pages={1855--1873}, year={2021}, doi={10.1109/TRO.2021.3075365}}

@article{fragapaneplanning2021, author={Fragapane, G. and {de Koster}, R. and Sgarbossa, F. and Strandhagen, J. O.}, title={Planning and control of autonomous mobile robots for intralogistics: Literature review and research agenda}, journal={European Journal of Operational Research}, volume={294}, number={2}, pages={405--426}, year={2021}, doi={10.1016/j.ejor.2021.01.019}}

@article{leanh2006, author={Le-Anh, T. and {de Koster}, M. B. M.}, title={A review of design and control of automated guided vehicle systems}, journal={European Journal of Operational Research}, volume={171}, number={1}, pages={1--23}, year={2006}, doi={10.1016/j.ejor.2005.01.036}}

@article{berbeglia2010, author={Berbeglia, G. and Cordeau, J.-F. and Laporte, G.}, title={Dynamic pickup and delivery problems}, journal={European Journal of Operational Research}, volume={202}, number={1}, pages={8--15}, year={2010}, doi={10.1016/j.ejor.2009.04.024}}

@article{pillac2013, author={Pillac, V. and Gendreau, M. and Gu{\'e}ret, C. and Medaglia, A. L.}, title={A review of dynamic vehicle routing problems}, journal={European Journal of Operational Research}, volume={225}, number={1}, pages={1--11}, year={2013}, doi={10.1016/j.ejor.2012.08.015}}

@article{hanne2009, author={Hanne, T. and Melo, T. and Nickel, S.}, title={Bringing robustness to patient flow management through optimized patient transports in hospitals}, journal={Interfaces}, volume={39}, number={3}, pages={241--255}, year={2009}, doi={10.1287/inte.1080.0379}}

@article{beaudry2010, author={Beaudry, A. and Laporte, G. and Melo, T. and Nickel, S.}, title={Dynamic transportation of patients in hospitals}, journal={OR Spectrum}, volume={32}, number={1}, pages={77--107}, year={2010}, doi={10.1007/s00291-008-0135-6}}

@article{kergosien2011, author={Kergosien, Y. and Lent{\'e}, C. and Piton, D. and Billaut, J.-C.}, title={A tabu search heuristic for the dynamic transportation of patients between care units}, journal={European Journal of Operational Research}, volume={214}, number={2}, pages={442--452}, year={2011}, doi={10.1016/j.ejor.2011.04.033}}

@article{bertsekas1997, author={Bertsekas, D. P. and Tsitsiklis, J. N. and Wu, C.}, title={Rollout algorithms for combinatorial optimization}, journal={Journal of Heuristics}, volume={3}, number={3}, pages={245--262}, year={1997}, doi={10.1023/A:1009635226865}}

@book{powell2011, author={Powell, W. B.}, title={Approximate Dynamic Programming: Solving the Curses of Dimensionality}, edition={2nd}, publisher={Wiley}, address={Hoboken, NJ}, year={2011}, doi={10.1002/9781118029176}}

@article{goodson2017, author={Goodson, J. C. and Thomas, B. W. and Ohlmann, J. W.}, title={A rollout algorithm framework for heuristic solutions to finite-horizon stochastic dynamic programs}, journal={European Journal of Operational Research}, volume={258}, number={1}, pages={216--229}, year={2017}, doi={10.1016/j.ejor.2016.09.040}}

@article{ulmer2019, author={Ulmer, M. W. and Goodson, J. C. and Mattfeld, D. C. and Hennig, M.}, title={Offline--online approximate dynamic programming for dynamic vehicle routing with stochastic requests}, journal={Transportation Science}, volume={53}, number={1}, pages={185--202}, year={2019}, doi={10.1287/trsc.2017.0767}}

@inproceedings{perron2023, author={Perron, L. and Didier, F. and Gay, S.}, title={The {CP-SAT-LP} Solver}, booktitle={29th International Conference on Principles and Practice of Constraint Programming (CP 2023)}, series={Leibniz International Proceedings in Informatics}, volume={280}, note={Article 3}, year={2023}, publisher={Schloss Dagstuhl}, address={Dagstuhl, Germany}, doi={10.4230/LIPIcs.CP.2023.3}}

@misc{ortools, author={Perron, L. and Furnon, V.}, title={{OR-Tools}}, howpublished={Google, \url{https://developers.google.com/optimization/}}, note={Version 9.15}, year={2026}}

@misc{hess2026, author={Hess, A. and Stampa, D.}, title={{Healthcare Logistics Dataset: GraphML Pathway Network and Transport Requests (University Medical Center Freiburg)}}, howpublished={Zenodo, version 1.0.0, \url{https://doi.org/10.5281/zenodo.19085901}}, year={2026}, doi={10.5281/zenodo.19085901}}

@article{monks2019, author={Monks, T. and Currie, C. S. M. and Onggo, B. S. and Robinson, S. and Kunc, M. and Taylor, S. J. E.}, title={Strengthening the reporting of empirical simulation studies: Introducing the {STRESS} guidelines}, journal={Journal of Simulation}, volume={13}, number={1}, pages={55--67}, year={2019}, doi={10.1080/17477778.2018.1442155}}

@article{sargent2013, author={Sargent, R. G.}, title={Verification and validation of simulation models}, journal={Journal of Simulation}, volume={7}, number={1}, pages={12--24}, year={2013}, doi={10.1057/jos.2012.20}}

@article{barton2014, author={Barton, R. R. and Nelson, B. L. and Xie, W.}, title={Quantifying input uncertainty via simulation confidence intervals}, journal={INFORMS Journal on Computing}, volume={26}, number={1}, pages={74--87}, year={2014}, doi={10.1287/ijoc.2013.0548}}

@article{holm1979, author={Holm, S.}, title={A simple sequentially rejective multiple test procedure}, journal={Scandinavian Journal of Statistics}, volume={6}, number={2}, pages={65--70}, year={1979}}

@book{imbens2015, author={Imbens, G. W. and Rubin, D. B.}, title={Causal Inference for Statistics, Social, and Biomedical Sciences: An Introduction}, publisher={Cambridge University Press}, address={Cambridge}, year={2015}, doi={10.1017/CBO9781139025751}}

@misc{siwek2026arxiv, author={Siwek, K. and {\'S}wietlicka, A.}, title={A State Based Dispatch Controller for Hospital Delivery Robots with Shared Human and Infrastructure Resources}, howpublished={arXiv preprint \href{https://arxiv.org/abs/2610.05971}{arXiv:2610.05971} [cs.RO]}, eprint={2610.05971}, archivePrefix={arXiv}, primaryClass={cs.RO}, year={2026}, doi={10.48550/arXiv.2610.05971}}

@article{ohneberg2025, author={Ohneberg, C. and St{\"o}bich, N. and Eberl, I.}, title={Possible Applications and Conditions for the Development and Integration of Robotic Assistance Systems in the Acute Inpatient Setting from the Perspective of Nursing Professionals: A Qualitative Study}, journal={International Journal of Social Robotics}, volume={17}, number={9}, pages={1821--1835}, year={2025}, doi={10.1007/s12369-025-01289-4}}

@article{brunett2025, author={Br{\"u}nett, M. and Gebert, A. and Weidner, F.}, title={Exploring Healthcare Staff's Attitudes and Perceptions of Robot Use in Nursing Care in {Germany}}, journal={International Journal of Social Robotics}, volume={17}, number={9}, pages={1635--1646}, year={2025}, doi={10.1007/s12369-025-01275-w}}

@article{klebbe2025, author={Klebbe, R. and Friese, C.}, title={Exploring the Role of Robots in Inpatient Care: Caregivers' Perspectives on the Development and Evaluation of Collaborative Robot Applications -- A Qualitative Research Approach}, journal={International Journal of Social Robotics}, volume={17}, number={9}, pages={1707--1724}, year={2025}, doi={10.1007/s12369-025-01270-1}}

@article{warmbein2025, author={Warmbein, A. and Sehn, L. and Rathgeber, I. and Seif, J. and Ohneberg, C. and St{\"o}bich, N. and Delker, A. and Zach, C. and Eberl, I. and Fischer, U.}, title={First Integration of a Service Robot and a Communication Application into a Nursing Isolation Setting -- An Observational Study Evaluating Walking Distances, Stress and Radiation Doses}, journal={International Journal of Social Robotics}, volume={17}, number={9}, pages={1809--1820}, year={2025}, doi={10.1007/s12369-025-01215-8}}

@article{faibish2022, author={Faibish, T. and Kshirsagar, A. and Hoffman, G. and Edan, Y.}, title={Human Preferences for Robot Eye Gaze in Human-to-Robot Handovers}, journal={International Journal of Social Robotics}, volume={14}, number={4}, pages={995--1012}, year={2022}, doi={10.1007/s12369-021-00836-z}}

@article{kimoto2022, author={Kimoto, M. and Yasumatsu, Y. and Imai, M.}, title={Help-Estimator: Robot Requests for Help from Humans by Estimating a Person's Subjective Time}, journal={International Journal of Social Robotics}, volume={14}, number={3}, pages={617--630}, year={2022}, doi={10.1007/s12369-021-00792-8}}

@article{hebesberger2017, author={Hebesberger, D. and Koertner, T. and Gisinger, C. and Pripfl, J.}, title={A Long-Term Autonomous Robot at a Care Hospital: A Mixed Methods Study on Social Acceptance and Experiences of Staff and Older Adults}, journal={International Journal of Social Robotics}, volume={9}, number={3}, pages={417--429}, year={2017}, doi={10.1007/s12369-016-0391-6}}

@article{turja2019, author={Turja, T. and Oksanen, A.}, title={Robot Acceptance at Work: A Multilevel Analysis Based on 27 {EU} Countries}, journal={International Journal of Social Robotics}, volume={11}, number={4}, pages={679--689}, year={2019}, doi={10.1007/s12369-019-00526-x}}

@article{rosenthal2012, author={Rosenthal, S. and Veloso, M. and Dey, A. K.}, title={Is Someone in this Office Available to Help Me?}, journal={Journal of Intelligent and Robotic Systems}, volume={66}, number={1--2}, pages={205--221}, year={2012}, doi={10.1007/s10846-011-9610-4}}

@inproceedings{srinivasan2016, author={Srinivasan, V. and Takayama, L.}, title={Help Me Please: Robot Politeness Strategies for Soliciting Help From Humans}, booktitle={Proceedings of the 2016 CHI Conference on Human Factors in Computing Systems}, pages={4945--4955}, publisher={ACM}, address={New York}, year={2016}, doi={10.1145/2858036.2858217}}

@article{gombolay2015, author={Gombolay, M. C. and Gutierrez, R. A. and Clarke, S. G. and Sturla, G. F. and Shah, J. A.}, title={Decision-making authority, team efficiency and human worker satisfaction in mixed human--robot teams}, journal={Autonomous Robots}, volume={39}, number={3}, pages={293--312}, year={2015}, doi={10.1007/s10514-015-9457-9}}

@article{hoffman2019, author={Hoffman, G.}, title={Evaluating Fluency in Human--Robot Collaboration}, journal={IEEE Transactions on Human-Machine Systems}, volume={49}, number={3}, pages={209--218}, year={2019}, doi={10.1109/THMS.2019.2904558}}

@article{mcfarlane2002, author={McFarlane, D. C. and Latorella, K. A.}, title={The Scope and Importance of Human Interruption in {Human--Computer} Interaction Design}, journal={Human--Computer Interaction}, volume={17}, number={1}, pages={1--61}, year={2002}, doi={10.1207/S15327051HCI1701_1}}

@article{westbrook2010, author={Westbrook, J. I. and Woods, A. and Rob, M. I. and Dunsmuir, W. T. M. and Day, R. O.}, title={Association of Interruptions With an Increased Risk and Severity of Medication Administration Errors}, journal={Archives of Internal Medicine}, volume={170}, number={8}, pages={683--690}, year={2010}, doi={10.1001/archinternmed.2010.65}}

@article{cvach2012, author={Cvach, M.}, title={Monitor Alarm Fatigue: An Integrative Review}, journal={Biomedical Instrumentation \& Technology}, volume={46}, number={4}, pages={268--277}, year={2012}, doi={10.2345/0899-8205-46.4.268}}

@article{lee2004, author={Lee, J. D. and See, K. A.}, title={Trust in Automation: Designing for Appropriate Reliance}, journal={Human Factors}, volume={46}, number={1}, pages={50--80}, year={2004}, doi={10.1518/hfes.46.1.50_30392}}
\end{document}